\documentclass{article}
\usepackage{fullpage,appendix}
\usepackage{cite}

\usepackage{amsmath,amsfonts,bm}

\def\eqref#1{equation~\ref{#1}}

\def\1{\bm{1}}

\DeclareMathAlphabet{\mathsfit}{\encodingdefault}{\sfdefault}{m}{sl}
\SetMathAlphabet{\mathsfit}{bold}{\encodingdefault}{\sfdefault}{bx}{n}

\usepackage{enumitem}
\usepackage[hidelinks]{hyperref}
\usepackage{algorithm}
\usepackage{algorithmic}
\usepackage{multirow}
\usepackage{multicol}
\usepackage{booktabs}
\usepackage{array}
\usepackage{xcolor}
\usepackage{amsmath,amsfonts,amssymb,amsthm}

\newtheorem{theorem}{Theorem}

\newtheorem{example}{Example}

\usepackage{graphicx} 
\usepackage[round]{natbib}

\usepackage{natbib} 
    \renewcommand{\bibsection}{\subsubsection*{References}}
\usepackage{mathtools} 
\usepackage{booktabs} 
\usepackage{tikz} 

\title{Sparse Domain Transfer via Elastic Net Regularization}

\author{
Jingwei~Zhang\thanks{Department of Computer Science and Engineering, The Chinese University of Hong Kong, \url{jwzhang22@ cse.cuhk.edu.hk}}~,
Farzan~Farnia\thanks{Department of Computer Science and Engineering, The Chinese University of Hong Kong, \url{farnia@cse.cuhk.edu.hk}}
	}   

\date{}
  
\begin{document}
\maketitle

\newcommand{\fix}{\marginpar{FIX}}
\newcommand{\new}{\marginpar{NEW}}

\newcommand{\anthon}{\textcolor{blue}}


\begin{abstract}
Transportation of samples across different domains is a central task in several machine learning problems. A sensible requirement for domain transfer tasks in computer vision and language domains is the sparsity of the transportation map, i.e., the transfer algorithm aims to modify the least number of input features while transporting samples across the source and target domains. 
In this work, we propose \emph{Elastic Net Optimal Transport (ENOT)} to address the sparse distribution transfer problem. The ENOT framework utilizes the $L_1$-norm and $L_2$-norm regularization mechanisms to find a sparse and stable transportation map between the source and target domains. To compute the ENOT transport map, we consider the dual formulation of the ENOT optimization task and prove that the sparsified gradient of the optimal potential function in the ENOT's dual representation provides the ENOT transport map. Furthermore, we demonstrate the application of the ENOT framework to perform feature selection for sparse domain transfer. We present the numerical results of applying ENOT to several domain transfer problems for synthetic Gaussian mixtures and real image and text data. Our empirical results indicate the success of the ENOT framework in identifying a sparse domain transport map.  
\end{abstract}

\section{Introduction}
Deep neural networks (DNNs) have revolutionized the performance of computer vision models in domain transfer applications where the features of an input sample are altered to transfer the sample to a secondary domain \citep{zhu2020sean, jiang2020tsit, ojha2021few}. The common goal of domain transfer algorithms is to transport an input data point to a target distribution by applying \emph{minimal changes} to the input. Over recent years, domain transportation algorithms based on generative adversarial networks (GANs) including CycleGAN \citep{zhu2017unpaired} and StyleGAN \citep{karras2019style} have achieved empirical success in addressing the domain transfer task for image distributions. The success of these algorithms has inspired several studies of GAN-based domain transfer methodologies \citep{zhao2020aclgan, chen2020reusing, van2019reversible}.

While the GAN-based methods have led to successful results in image-based domain transfer problems, their application demands significantly higher computational costs than standard GAN algorithms including only one generator/discriminator neural net pair to transfer a latent Gaussian vector to the data distribution. The extra computations in these domain transfer algorithms aim to ensure an invertible transfer map and thus limited modifications to an input sample.  For example, the CycleGAN algorithm considers two pairs of generator/discriminator neural nets to impose a reversible transformation of an input image. However, the additional pair of neural nets in the CycleGAN setting will lead to a more challenging optimization task and higher training costs. 

In this work, we focus on \emph{sparse domain transfer problems} where the transfer of samples between source and target domains can be achieved by editing only a limited subset of input features. We note that the assumption of 
 a sparse transport map applies to several real-world domain transfer problems, e.g. object translation, text revision, and gene editing problems. In the mentioned tasks, the sparsity level of the transfer map results in a meaningful measure of changes applied to an input sample. While sparse transportation maps are desired in many real-world domain transfer problems, the commonly-used GAN-based algorithms often lead to dense transfer maps editing a considerable fraction of input features. 

To address sparse domain transfer tasks, we propose an optimal transport-based approach which takes advantage of the induced sparsity of the $L_1$-norm regularization and the stability properties of the $L_2$-norm regularization. Our proposed framework, which we call \textit{Elastic Net Optimal Transport (ENOT)}, solves an optimal transport problem where the transportation cost follows from the elastic net function \citep{zou2005regularization} combining standard Euclidean-norm-squared and $L_1$-norm cost functions. Therefore, the ENOT approach can be interpreted as a mechanism to regularize the standard optimal transport map toward sparser transportation functions. By tuning the coefficient of the $L_1$-norm regularization in ENOT's elastic net cost, the learner can adjust the sparsity level of the transportation map and explore the spectrum between the standard and fully $L_1$-norm-based optimal transport tasks.

To analyze the ENOT problem, we leverage optimal transport theory \citep{villani2009optimal} and extend the duality results to the ENOT optimal transport setting. We prove a generalization of standard Brenier's theorem, highlighting the connection between the optimal potential function in the ENOT's dual problem with the optimal transport map transferring samples across domains. Our main theorem suggests that the composition of a soft-thresholding function with the gradient of the optimal potential function will perform sparse transportation across the domains.  This result indicates that the ENOT framework offers a combination of the standard optimal transport problem with the squared-error cost function and the $L_1$-norm-based optimal transport problem which leads to a challenging optimization problem without the $L_2$-norm-based regularization in ENOT. 

Furthermore, we utilize the ENOT framework to develop a \textit{feature selection-based approach} to reduce the sparse domain transport task to a constraint-free distribution transfer problem where an unconstrained transfer map is applied to only the selected feature subset. According to this variable selection-based approach, we break the domain transfer problem into two sub-problems: 1) ENOT-based variable selection choosing features to undergo modification for a given input sample, 2) 
Applying an unconstrained transportation map via standard GAN frameworks that transfers the input sample masked by the feature selection output to the target domain. 
By tuning the coefficient of the $L_1$-norm regularization in the ENOT's elastic net cost, the learner can adjust the number of selected features prior to performing a constraint-free GAN-based distribution transfer. 


Finally, we discuss the numerical results of the applications of the ENOT framework to sparse domain transfer problems from various areas including computer vision, computational biology, and natural language processing. Our empirical findings show that the feature selection-based domain transfer via ENOT can be easily adapted to different domains and achieves satisfactory results. We qualitatively evaluate ENOT's application for feature identification in sparse domain transfer. The numerical results support the proposed methodology of sparse domain transfer via ENOT-based domain transportation and feature selection. The contributions of this work can be summarized as:

\begin{itemize}[leftmargin=*]
    \item Proposing a feature selection-based approach for the  sparse domain transfer problem,
    \item Developing ENOT as an elastic net-based methodology to the sparse domain transfer and variable selection,
    \item Extending the theory of standard squared-error-based optimal transport task to the ENOT setting,
    \item Providing supportive numerical results for applying ENOT-based sparse domain transfer to various domain transfer tasks.
\end{itemize}

\section{Related Work}
\textbf{Sparsity and Optimal Transport Methods.} Several related works have studied various notions of sparsity in optimal transport frameworks. 
References \citep{flamary2016optimal, blondel2018smooth,swanson2020rationalizing,bao2022sparse,liu2023sparsityconstrained} propose sparsity-based regularization of the transportation matrix in optimal transport problems. 
However, we note that the sparsity objective pursued in these works differs from the sparse domain transfer in our work: while the mentioned papers aim for a sparse transportation matrix to gain a sparse alignment of source and target samples, our proposed ENOT method focuses on the sparsity of the modified input features in the domain transfer.  

\cite{cuturi2023monge} first introduces an optimal transport-based approach to the sparse domain transfer problem, which aim to precisely solve an entropic-regularized optimal transport problem over the empirical samples and then use a kernel-based interpolation to generalize the solution to unseen data. On the other hand, our approach to the optimal transport problem is different and more similar to the Wasserstein GAN framework, where we leverage a neural net potential function which can be extended to large-scale image and text data. Overall, our neural net-based method for the ENOT optimal transport problem can be viewed as a complementary approach to the precise kernel-based framework in \citep{cuturi2023monge}. Also, the trained neural net can be used as an efficient-to-compute feature selection map, which we later used to introduce a \textit{feature selection-based approach} to sparse domain transfer, a topic that has not been studied in \citep{cuturi2023monge}, which is useful for large-scale image and text-related applications.

\textbf{Unsupervised Image to Image Translation (UI2I).} 
Several related works attempt to address image-based transportation problems. For the image style transfer task, CycleGAN \citep{zhu2017unpaired} uses a cycle-consistent loss and two GANs to conduct cyclic unpaired transformation. DRIT++ \citep{lee2020drit++} adopts encoders to obtain the latent representation of images and similar cross-cycle consistency loss. For the image colorization transfer task, Conditional GANs are leveraged to improve colorization performance \citep{isola2017image}. \cite{zhang2017real} propose a real-time user-guided neural network colorization. Moreover, cyclic-loss \citep{bhattacharjee2020dunit, wu2019transgaga, shen2019towards,zhu2017unpaired} and GANs \citep{zhao2020aclgan, chen2020reusing, van2019reversible} have been utilized to address UI2I. However, unlike ENOT, these related works do not focus on the sparsity of transfer maps. 

\textbf{Sequence to Sequence Translation. }
Sequence to sequence (Seq2Seq) neural net models are typically designed based on an encoder-decoder architecture. \cite{kalchbrenner2013recurrent} propose the application of a convolutional neural network (CNN) as the encoder and a recurrent neural network (RNN) as the decoder. \cite{sutskever2014sequence} utilize an RNN-based architecture for both the encoder and decoder neural nets. \cite{vaswani2017attention} propose a transformer based on multi-head self-attention. BART \citep{lewis2019bart} offers a sequence-to-sequence pretraining solution and adopts a bidirectional encoder similar to BERT \citep{devlin2018bert}, and a decoder similar to GPT \citep{radford2018improving}. Unlike our proposed ENOT approach, the discussed methods usually result in a dense transportation map. We also note that Seq2Seq and GAN-based transfer methods are almost exclusively used for language and image distributions, respectively. 

\section{Preliminaries}
Consider random vectors $\mathbf{X},\mathbf{Y} \in \mathbb{R}^d$ with probability distributions $P_X,P_Y$, respectively. Given $n$ independent samples $\mathbf{x}_1,\ldots,\mathbf{x}_n$ from $P_\mathbf{X}$ and $m$ independent samples $\mathbf{y}_1,\ldots,\mathbf{y}_m$  from $P_\mathbf{Y}$, the goal in the domain transfer problem is to learn a map $\psi:\mathbb{R}^d \rightarrow \mathbb{R}^d$ transporting an input $\mathbf{X}$ from distribution $P_X$ to an output $\psi(\mathbf{X})$ distributed as $P_Y$, i.e.,
\begin{equation*}
    \psi(\mathbf{X}) \: \stackrel{\text{\rm dist}}{=} \: \mathbf{Y}.
\end{equation*}
Here, $\stackrel{\text{\rm dist}}{=}$ denotes identical probability distributions. 

Without any constraint on the map $\psi$, there exist infinitely many transportation maps resulting in the required identical distributions. To uniquely characterize the transfer map, the optimal transport framework \citep{villani2009optimal} seeks to find a map minimizing the expected transportation cost measured based on a cost function $c:\mathbb{R}^d\times \mathbb{R}^d\rightarrow \mathbb{R}$. According to this framework, the transportation map follows from the optimal coupling $\Pi_{X,Y}$, marginally distributed as $P_{X}$ and $P_Y$, that is minimizing the expected transportation cost formulated as
\begin{equation*}
    \mathrm{OT}_c(P_X,P_Y):= \underset{\substack{\Pi_{X,Y}:\, \Pi_{X} = P_X \\ \qquad\;\; \Pi_{Y} = P_Y} }{\inf}\; \mathbb{E}_{(X,Y)\sim\Pi}\bigl[c(\mathbf{X},\mathbf{Y}) \bigr].
\end{equation*}
Here, $ \mathrm{OT}_c(P_X,P_Y)$ denotes the optimal transport cost between $P_X,P_Y$.
It is well-known that under mild regularity conditions, a deterministic coupling mapping  $\mathbf{X}$ to a sample with distribution $P_Y$ exists that solves the above problem. Also, the dual representation of the above optimization problem can be formulated via the Kantorovich duality \citep{villani2009optimal} as
\begin{equation*}  \sup_{\phi:\mathbb{R}^d\rightarrow\mathbb{R}}\; \mathbb{E}\bigl[ \phi(\mathbf{X})\bigr] - \mathbb{E}\bigl[ \phi^c(\mathbf{Y})\bigr],
\end{equation*}
where $\phi$ is the potential function and the $c$-transform $\phi^c$ is defined as $\phi^c(\mathbf{y}):= \sup_{\mathbf{y}'} \phi(\mathbf{y}') - c(\mathbf{y},\mathbf{y}')$. 

\begin{example}
In the special case of a norm cost $c_1(\mathbf{x},\mathbf{y}) = \Vert \mathbf{x} - \mathbf{y}\Vert$, the result of Kantorovich duality can be written as 
\begin{equation*}
 \mathrm{OT}_{c_1}(P_X,P_Y) \, =\, \sup_{\phi: \text{\rm 1-Lipschitz }}\; \mathbb{E}\bigl[ \phi(\mathbf{X})\bigr] - \mathbb{E}\bigl[ \phi(\mathbf{Y})\bigr]   
\end{equation*}
where the potential function $\phi:\mathbb{R}^d\rightarrow\mathbb{R}$ is constrained to be $1$-Lipschitz with respect to the assigned norm $\Vert \cdot \Vert$, i.e., for every $\mathbf{x},\mathbf{x}'\in \mathbb{R}^d$:
\begin{equation*}
    \bigl\vert \phi(\mathbf{x}) - \phi(\mathbf{x}') \bigr\vert \, \le\, \bigl\Vert \mathbf{x}- \mathbf{x}'\bigr\Vert.
\end{equation*}
\end{example}

\begin{example}\label{Example: OT L2squared}
In the special case of the $L_2$-norm-squared cost $c_2(\mathbf{x},\mathbf{y}) = \frac{1}{2}\Vert \mathbf{x} - \mathbf{y}\Vert^2_2$, the result of Kantorovich duality can be written as 
\begin{equation}\label{Eq: OT 2-Wasserstein dual}
  \sup_{\widetilde{\phi}: \text{\rm convex }}\; \mathbb{E}\Bigl[\frac{1}{2}\Vert \mathbf{X}\Vert^2_2 - \widetilde{\phi}(\mathbf{X})\Bigr] + \mathbb{E}\Bigl[\frac{1}{2}\Vert \mathbf{Y}\Vert^2_2 - \widetilde{\phi}^\star(\mathbf{Y})\Bigr]   
\end{equation}
where the potential function $\phi(\mathbf{x}) := \frac{1}{2}\Vert \mathbf{x}\Vert^2_2 - \widetilde{\phi}(\mathbf{x}) $ is constrained to be the subtraction of a convex function $\widetilde{\phi}$ from $\frac{1}{2}\Vert \mathbf{x}\Vert^2_2$, and $\widetilde{\phi}^\star$ is the Fenchel conjugate defined as
\begin{equation*}
    \widetilde{\phi}^\star(\mathbf{x}) \, := \, \sup_{\mathbf{x}'} \mathbf{x}'^\top \mathbf{x} - \widetilde{\phi}(\mathbf{x}').
\end{equation*}
\end{example}
The Brenier theorem reveals that in the setting of Example~\ref{Example: OT L2squared}, the gradient of the optimal solution $\widetilde{\phi}$ provides the unique monotone (gradient of a convex function) map transporting samples between the two domains:
\begin{theorem}[Brenier's Theorem, \citep{villani2009optimal}]
Suppose that $P_X,P_Y$ are absolutely continuous with respect to one another. Then, the gradient of the solution $\widetilde{\phi}^*$ to \eqref{Eq: OT 2-Wasserstein dual} is the unique monotone map for transferring $P_X$ to $P_Y$, that is
\begin{equation*}
    \nabla \widetilde{\phi}^* (\mathbf{X}) \: \stackrel{\text{\rm dist}}{=} \: \mathbf{Y}.
\end{equation*}
\end{theorem}

In the following sections, we aim to define and analyze optimal transport costs that can capture the sparsity of the transportation map, i.e. the number of non-zero coordinates of $\mathbf{y}-\mathbf{x}$.
\section{Elastic Net Regularization for Sparse Optimal Transport}
In this work, we aim to address the sparse domain transfer problem where the transfer map $\psi$ between distributions $P_X,P_Y$ alters the fewest possible coordinates in the 
 $d$-dimensional feature vector $\mathbf{X}=[X^{(1)},\ldots,X^{(d)}]$. To apply the optimal transport framework, a proper cost function is the cardinality (number of non-zero elements $\mathrm{card}(\mathbf{z})=\sum_{i=1}^d \mathbf{1}[z_i\neq 0]$) of the difference between the original and transported samples:
 \begin{equation*}
     c_{\mathrm{sparse}}(\mathbf{x},\mathbf{y}) = \mathrm{card}(\mathbf{x}-\mathbf{y}).
 \end{equation*}
 Since the cardinality function lacks continuity and convexity, the resulting optimal transport problem will be computationally difficult. A common convex proxy for the cardinalty function is the $L_1$-norm where we simply use $c_{L_1}(\mathbf{x},\mathbf{y}) =\Vert \mathbf{x}-\mathbf{y}\Vert_1$. While the primal optimal transport problem could be solved for the empirical samples with the $L_1$-norm cost, the domain transfer map requires solving the optimization problem for the data distribution which would be complex in the primal case.
 Therefore, we focus on the dual optimization problem to the optimal transport task. However, solving the dual optimization problem of the $L_1$-norm cost requires optimizing over the $L_1$-norm-based 1-Lipschitz functions which would be challenging.  

 To handle the computational complexity of the dual optimization problem with $L_1$-norm cost function, we propose to apply the \emph{elastic net} \citep{zou2005regularization} cost function with coefficients $0\le \alpha\le 1$ and $\lambda >0$:
\begin{equation}\label{Eq: EN_cost}
    c^{\alpha,\lambda}_{\mathrm{EN}}(\mathbf{x},\mathbf{y}) = \lambda (1-\alpha) \bigl\Vert \mathbf{x}-\mathbf{y} \bigr\Vert^2_2 + \lambda  \alpha \bigl\Vert \mathbf{x}-\mathbf{y} \bigr\Vert_1.
\end{equation}
Using the above cost function, we propose the \emph{Elastic Net-based Optimal Transport (ENOT)} as the optimal transport method formulated with the cost function in \eqref{Eq: EN_cost}. For the dual formulation of the ENOT problem, we can apply the Kantorovich duality to obtain the following optimization task:
\begin{equation}\label{Eq: ENOT dual problem}
    \max_{\phi:\mathbb{R}^d\rightarrow\mathbb{R}}\: \mathbb{E}\bigl[\phi(\mathbf{X})\bigr] - \mathbb{E}\bigl[\phi^{c^{\alpha,\lambda}_{\mathrm{EN}}}(\mathbf{Y})\bigr] 
\end{equation}
where the elastic-net-based $c$-transform can be written as follows:
\begin{equation*}
    \phi^{c^{\alpha,\lambda}_{\mathrm{EN}}}(\mathbf{y}) := \max_{\boldsymbol{\delta}\in\mathbb{R}^d}\; \phi(\mathbf{y} + \boldsymbol{\delta} ) -  \lambda (1-\alpha) \Vert \boldsymbol{\delta} \Vert^2_2 - \lambda  \alpha \Vert \boldsymbol{\delta} \Vert_1.
\end{equation*}

\begin{theorem}\label{Thm: Weak Concavity: ENOT potential}
    Consider the ENOT dual problem in \eqref{Eq: ENOT dual problem}. Then, there exists an optimal potential function $\phi^*$ for this problem 
    which satisfies the following \textit{weakly-concavity} property: for every $\mathbf{x,y}\in\mathbb{R}^d$ and real value $ \gamma\in [0,1]$:  
\begin{equation*}
\begin{aligned}
&\phi^*\bigl(\gamma\mathbf{x} + (1-\gamma)\mathbf{y}\bigr)\, \ge\, \gamma \phi^*(\mathbf{x}) + (1-\gamma)\phi^*(\mathbf{y}) - \lambda\gamma(1-\gamma)(1-\alpha)\bigl\Vert\mathbf{x}-\mathbf{y} \bigr\Vert^2_2 - \lambda\alpha \bigl\Vert\mathbf{x}-\mathbf{y} \bigr\Vert_1 .
    \end{aligned}
\end{equation*}
\end{theorem}
\begin{proof}
    We defer the proof to the Appendix.
\end{proof}
The above result shows the existence of an optimal potential function possessing a weakly-concave structure defined based on an elastic net function. Our next result reveals the extension of the Brenier's theorem to the elastic net cost function. In this extension, we use $\mathrm{ST}_{\gamma}$ to denote the soft-thresholding operator defined for a scalar input as
\begin{equation*}
   \mathrm{ST}_{\gamma}(z) := \begin{cases}
   z+\gamma \; &\text{\rm if }\; z\le -\gamma \\ 
   0 \; &\text{\rm if }\; -\gamma< z< \gamma \\
   z-\gamma \; &\text{\rm if }\;  \gamma \le z.
   \end{cases} 
\end{equation*}
For a vector input $\mathbf{z}\in\mathbb{R}^d$, we define the soft-thresholding map as the coordinate-wise application of the scalar soft-thresholding function, i.e.,  $$\forall i\in\{1,\ldots ,d\}: \quad \mathrm{ST}_{\gamma}(\mathbf{z})_i = \mathrm{ST}_{\gamma}(z_i)$$ 
\begin{theorem}\label{Theorem: Brenier, ENOT}
Consider the dual ENOT problem in \eqref{Eq: ENOT dual problem}. Then, given the optimal potential function $\phi^*$ the following will provide the optimal transport map transferring samples across domains:
\begin{equation*}
    \mathbf{X} - \mathrm{ST}_{\frac{\alpha}{2(1-\alpha)}}\Bigl(\frac{1}{2\lambda(1-\alpha)}\nabla \phi^*(\mathbf{X})\Bigr) \, \stackrel{\mathrm{dist}}{=}\, \mathbf{Y}. 
\end{equation*}
\end{theorem}
\begin{proof}
    We defer the proof to the Appendix.
\end{proof}
Note that the above theorem is a generalization of the Brenier theorem for the elastic net cost, and in the special case of $\alpha=0$ reduces to the Brenier theorem. On the other hand, by selecting a larger $L_1$-regularization coefficient $\alpha$, the soft-thresholding map will apply a more stringent sparsification to the gradient map of the optimal potential function. This result suggests that by choosing a larger $\alpha$, one can achieve a sparser transportation map which is the goal sought by the sparse transfer algorithm. Theorem \ref{Theorem: Brenier, ENOT} reduces the search for the elastic net-based transport map to the computation of the optimal potential function $\phi$ in the dual optimization problem, which as shown in Theorem~\ref{Thm: Weak Concavity: ENOT potential} satisfies a weakly-concavity property. 
\section{ENOT-based Feature Selection for Sparse Domain Transfer}

\begin{table*}
  \caption{ENOT's achieved NLL with different coefficients of $L_1$-regularization on the Gaussian mixture transfer.}
  \label{table-gau-l1}
  \centering
  \begin{tabular}{lllllllll}
    \toprule
    & & \multicolumn{7}{c}{ENOT $L_1$ coefficient}                   \\
    \cmidrule(r){3-9}
    $f$  & {\small dimension} & baseline 0 & 1e-3 & 5e-3 & 1e-2 & 5e-2 & 1e-1 & 5e-1\\
    \midrule
    \multirow{3}{*}{MLP-22} & $1000$ & 4.87$\times 10^\mathbf{ 3}$ & 4.46$\times 10^\mathbf{ 3}$ & 4.13$\times 10^\mathbf{ 3}$ & 3.52$\times 10^\mathbf{ 3}$ & \bf1.50$\times 10^\mathbf{ 3}$ & 1.50$\times 10^\mathbf{3}$ & 1.51$\times 10^\mathbf{3}$\\
    & $100$ & 3.22$\times 10^\mathbf{ 3}$ & 3.15$\times 10^\mathbf{ 3}$ & 2.97$\times 10^\mathbf{ 3}$ & 1.82$\times 10^\mathbf{ 3}$ & \bf1.62$\times 10^\mathbf{ 2}$ & 1.64$\times 10^\mathbf{ 2}$ & 1.82$\times 10^\mathbf{ 2}$\\
    & $10$ & 1.53$\times 10^\mathbf{ 1}$ & 1.50$\times 10^\mathbf{ 1}$ & 1.42$\times 10^\mathbf{ 1}$ & 1.39$\times 10^\mathbf{ 1}$ & \bf1.31$\times 10^\mathbf{ 1}$ & 1.40$\times 10^\mathbf{ 1}$ & 1.51$\times 10^\mathbf{ 1}$\\
    \midrule
    \multirow{3}{*}{MLP-12}     & $1000$ & 4.56$\times 10^\mathbf{ 3}$ & 4.42$\times 10^\mathbf{ 3}$ & 4.02$\times 10^\mathbf{ 3}$ & 3.48$\times 10^\mathbf{ 3}$  & \bf1.50$\times 10^\mathbf{ 3}$  & 1.50$\times 10^\mathbf{ 3}$  & 1.50$\times 10^\mathbf{ 3}$   \\
    & $100$ & 2.73$\times 10^\mathbf{ 3}$ & 2.58$\times 10^\mathbf{ 3}$ & 2.85$\times 10^\mathbf{ 3}$ & 1.27$\times 10^\mathbf{ 3}$ & \bf1.50$\times 10^\mathbf{ 2}$ & 1.51$\times 10^\mathbf{ 2}$ & 1.51$\times 10^\mathbf{ 2}$\\
    & $10$ & 1.60$\times 10^\mathbf{ 1}$ & 1.39$\times 10^\mathbf{ 1}$ & 1.51$\times 10^\mathbf{ 1}$ & 1.48$\times 10^\mathbf{ 1}$ & \bf1.34$\times 10^\mathbf{ 1}$ & 1.48$\times 10^\mathbf{ 1}$ & 1.67$\times 10^\mathbf{ 1}$\\
    \midrule
    \multirow{3}{*}{MLP-4} & $1000$ & 3.67$\times 10^\mathbf{ 3}$ & 3.62$\times 10^\mathbf{ 3}$ & 3.44$\times 10^\mathbf{ 3}$ & 3.01$\times 10^\mathbf{ 3}$ & \bf1.50$\times 10^\mathbf{ 3}$ & 1.50$\times 10^\mathbf{ 3}$  & 1.50$\times 10^\mathbf{ 3}$ \\
    & $100$ & 2.62$\times 10^\mathbf{ 3}$ & 2.53$\times 10^\mathbf{ 3}$ & 1.44$\times 10^\mathbf{ 3}$ & 1.01$\times 10^\mathbf{ 3}$ & \bf1.50$\times 10^\mathbf{ 2}$ & 1.51$\times 10^\mathbf{ 2}$  & 1.51$\times 10^\mathbf{ 2}$\\
    & $10$ & 1.52$\times 10^\mathbf{ 1}$ & 1.36$\times 10^\mathbf{ 1}$  & 1.38$\times 10^\mathbf{ 1}$ & 1.44$\times 10^\mathbf{ 1}$  &\bf1.32$\times 10^\mathbf{ 1}$ & 1.47$\times 10^\mathbf{1}$  & 1.77$\times 10^\mathbf{1}$\\
    \bottomrule
  \end{tabular}
\end{table*}

In the previous section, we have shown sparse transfer map could be derived by applying the soft-thresholding function to the gradient of the optimal potential function. In addition to directly performing a sparse optimal transport, the trained potential function in the ENOT framework can be used for variable selection to undergo an unconstrained distribution transfer. Therefore, we also propose a feature selection algorithm for domain transfer using the optimal potential function $\phi^*$ in \eqref{Eq: ENOT dual problem}. Here for an input $\mathbf{x}\in\mathbb{R}^d$, we define the feature selection mask $I:\mathbb{R}^d\rightarrow \{0,1\}^d$ as
\begin{equation}\label{Eq: ENOT mask}
    \forall i\in\{1,\ldots ,d\}: I(\mathbf{x})_i= \begin{cases}
    0\quad &\text{\rm if}\; \bigl\vert \nabla\phi^*(\mathbf{x})_i\bigr\vert \le \lambda\alpha, \\
     1\quad &\text{\rm if}\; \bigl\vert \nabla\phi^*(\mathbf{x})_i\bigr\vert > \lambda\alpha
    \end{cases}
\end{equation}
The above masking identifies the feature coordinates modified by the ENOT transport map. Given the above masking function, we can train a generator function $G:\mathbb{R}^d\rightarrow\mathbb{R}^d$ to perform a constraint-free domain transportation on the ENOT's selected features. We can employ the standard GAN framework \citep{goodfellow2014generative} consisting of a generator $G$ and discriminator function $D:\mathbb{R}^d\rightarrow \mathbb{R}$ to do this task. Following the standard min-max formulation of GANs, we propose the following optimization problem for the ENOT feature selection-based domain transport:
\begin{align}
    &\min_{G\in\mathcal{G}}\; \max_{D\in\mathcal{D}}\;\mathbb{E}\Bigl[\log\Bigl(D(\mathbf{Y})\Bigr)\Bigr]+ \mathbb{E}\Bigl[\log\Bigl(1- D\Bigl(\, I(\mathbf{X})\odot G(\mathbf{X})  +  \bigl(1-I(\mathbf{X})\bigr)\odot\mathbf{X}\, \Bigr)\Bigr)\Bigr] \nonumber
\end{align}
In the above, the generator $G$ attempts to match the distribution of modified $I(\mathbf{X})\odot G(\mathbf{X})  +  \bigl(1-I(\mathbf{X})\bigr)\odot\mathbf{X}$ with the distribution of $\mathbf{Y}$,  where $\odot$ denotes the element-wise Hadamard product. On the other hand, the discriminator $D$ seeks to identify the original $\mathbf{Y}$ samples from the modified $\mathbf{X}$ data. Since we utilize the feature selection mask of the trained ENOT potential function, we do not need to ensure the invertibility of the generator and can reduce the number of machine players compared to the CycleGAN algorithm.

The above feature selection-based approach enables the application of neural net generator functions which could improve the vanilla ENOT's performance due to the power of a properly-designed generator to model the structures in the text and image data. This is similar to the Wasserstein GAN (WGAN) \citep{arjovsky2017wasserstein} as the optimal-transport-based GAN formulation in WGANs also considers a generator $G$ instead of relying on the gradient of the potential function. 


\section{Numerical Results}
\begin{figure*}
  \centering
  \includegraphics[width=.52\textwidth]{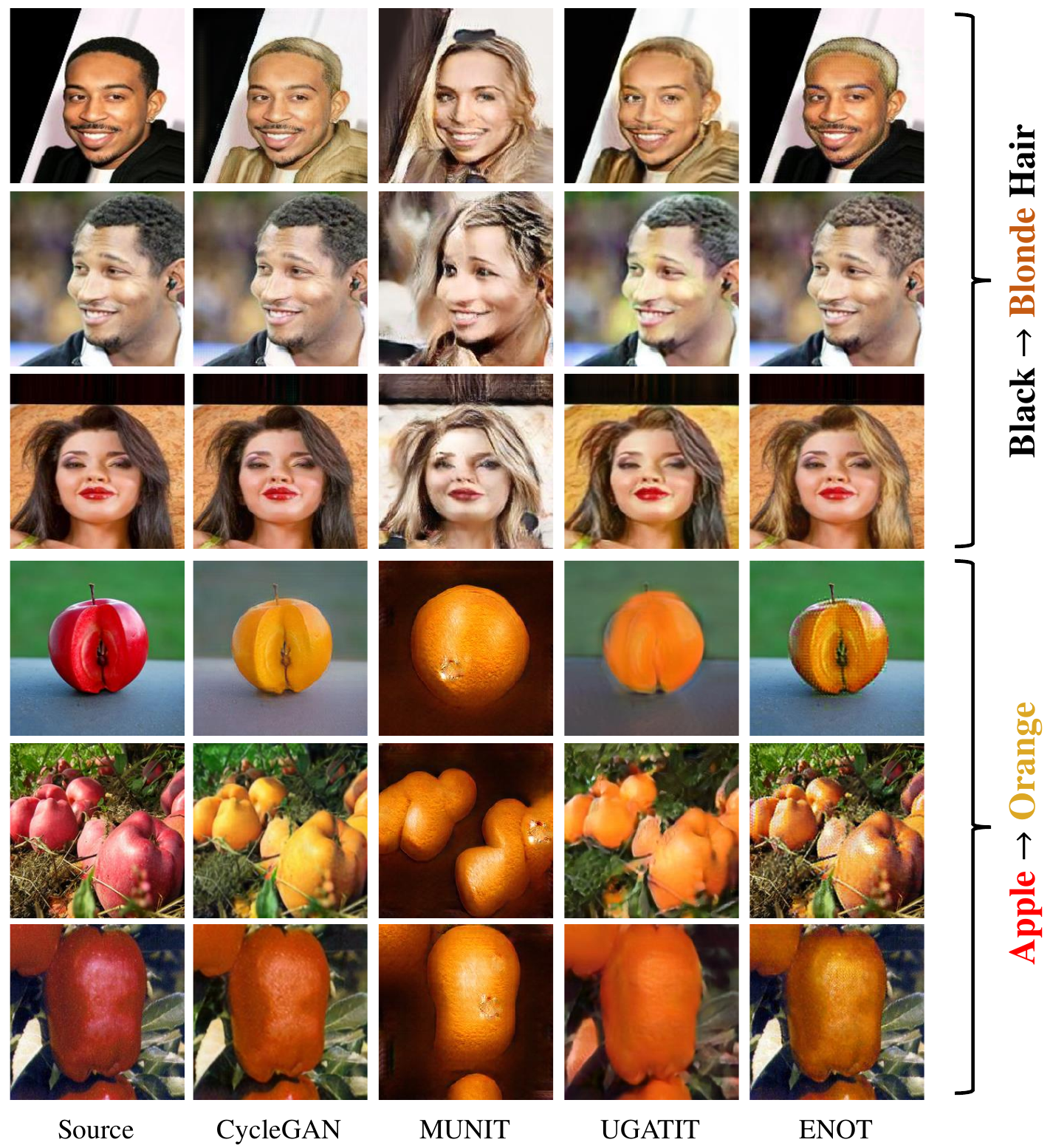}
  \caption{Transportation 
  for Black$\rightarrow$Blonde~hair and  Apple$\rightarrow$Orange on CelebA and Apple2Orange. }
  \label{fig:realimage}
\end{figure*}

\begin{figure*}
  \centering
  \includegraphics[width=.52\textwidth]{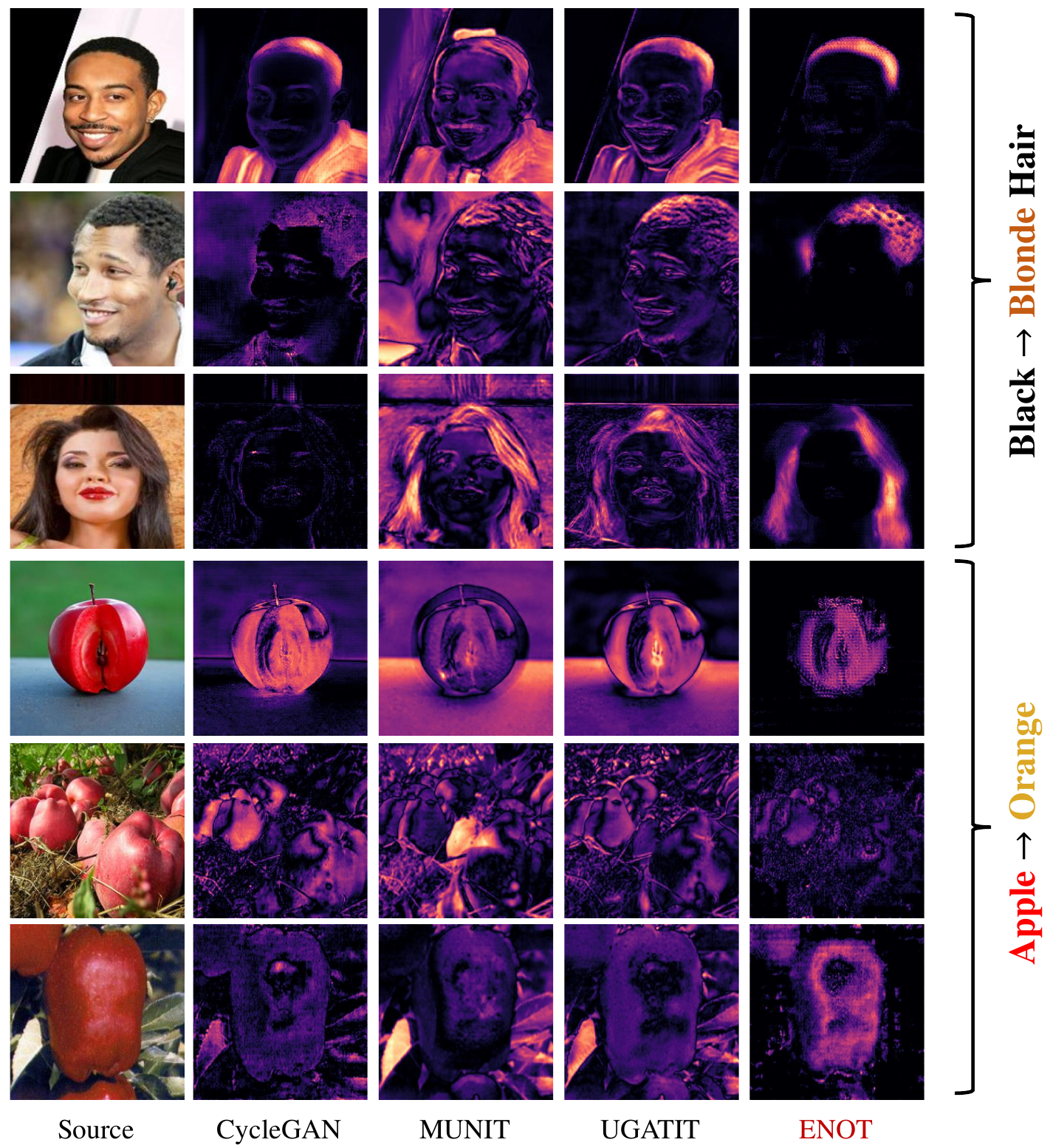}
  \caption{Saliency maps of transportation 
  for Black$\rightarrow$Blonde~hair and  Apple$\rightarrow$Orange on CelebA and Apple2Orange. }
  \label{fig:realimage_saliency}
\end{figure*}

In this section, we present the empirical results of the applications of ENOT and the baseline domain transfer algorithms to several standard datasets, including synthetic Gaussian mixture models, and real image and text datasets. We defer the details of our numerical experiments, including the dataset pre-processing, neural network architectures, and hyperparameter selection to the Appendix.


\subsection{ENOT applied to Synthetic Gaussian Mixture Data}


We evaluated the performance of ENOT in domain transfer problems across multivariate Gaussian mixture models (GMMs). In our experiments, we considered bimodal source and target GMMs: the source GMM  $p(\boldsymbol{x})= \phi_s \mathcal{N}\left(\boldsymbol{x}|\boldsymbol{\mu}_{\boldsymbol{s}}, \sigma^2\boldsymbol{ I}_{\boldsymbol{d}}\right)+\left(1-\phi_s\right) \mathcal{N}\left(\boldsymbol{x}|-\boldsymbol{\mu}_{\boldsymbol{s}},\sigma^2\boldsymbol{I}_{\boldsymbol{d}}\right)$, and target GMM $p(\boldsymbol{y})= \phi_t\mathcal{N}\left(\boldsymbol{y}|\boldsymbol{\mu}_{\boldsymbol{t}}, \sigma^2\boldsymbol{ I}_{\boldsymbol{d}}\right)+\left(1-\phi_t\right) \mathcal{N}\left(\boldsymbol{y}|-\boldsymbol{\mu}_{\boldsymbol{t}},\sigma^2\boldsymbol{I}_{\boldsymbol{d}}\right)$ both consist of two multivariate Gaussian components with different means and identical covariance matrix with $\sigma=1$. There exist two component-based mappings: (1) mapping $\mathcal{N}_{\boldsymbol{\mu}_{\boldsymbol{s}}} \rightarrow \mathcal{N}_{\boldsymbol{\mu}_{\boldsymbol{t}}},\, \mathcal{N}_{-\boldsymbol{\mu}_{\boldsymbol{s}}} \rightarrow \mathcal{N}_{-\boldsymbol{\mu}_{\boldsymbol{t}}}$ or (2) mapping $\mathcal{N}_{\boldsymbol{\mu}_{\boldsymbol{s}}} \rightarrow \mathcal{N}_{-\boldsymbol{\mu}_{\boldsymbol{t}}},\, \mathcal{N}_{-\boldsymbol{\mu}_{\boldsymbol{s}}} \rightarrow \mathcal{N}_{\boldsymbol{\mu}_{\boldsymbol{t}}}$. We set 
$\boldsymbol{\mu}_{\boldsymbol{s}} = \left[ \gamma, \epsilon_d, \cdots, \epsilon_d \right], \boldsymbol{\mu}_{\boldsymbol{t}} = \left[ -\gamma, \epsilon_d, \cdots, \epsilon_d\right]$ and chose $\epsilon_d \cdot \sqrt{d-1}<\gamma$ to distinguish the optimal $L_1$-norm-based sparse and standard $L_2$-norm-based transfer maps. We set $\gamma=10,\; \epsilon_{10}=2,\; \phi_s=\phi_t=0.5$, and scaled $\epsilon_d = \frac{\epsilon_{10}}{\sqrt{d/10}}$ to ensure the inequality holds in different dimensions. 


We applied the ENOT approach by solving the dual optimization problem (Eq. \ref{Eq: ENOT dual problem}) using a multi-layer perception neural net with different number of ReLU layers. We attempted different $L_1$-norm coefficients, where a zero coefficient reduces to the standard optimal transport baseline. We evaluated the performance of the domain transfer algorithm using the averaged negative log-likelihood (NLL) of transferred samples with respect to the target Gaussian mixture distribution. 
Based on our quantitative results in Table \ref{table-gau-l1}, we observed that the ENOT's sparse transfer maps led to better performance scores for the three potential function architectures. 


\begin{table*}[t]
  \caption{IMDB Movie Review Sentiment Transportation Maps}
  \label{table-imdb}
  \centering
  \begin{tabular}{p{0.3\textwidth}p{0.3\textwidth}p{0.3\textwidth}}
    \toprule
    \multicolumn{3}{c}{\centering{Transfer Task: Negative Review $\rightarrow$ Positive Review}}
     \\ 
     \multicolumn{3}{c}{\centering{\textcolor{red}{Red: ENOT's Selected Words for Domain Transfer.}
     \textcolor{blue}{Blue: Modified Parts}
     }}
     \\
    \midrule 
     Source & Baseline Seq2Seq & ENOT \\
    \midrule 
     
    Lonely, disconnected, middle-class housewife in the midst of a divorce seeks solace to reflect on her immediate future. at some sort of bed and breakfast by ( well, literally in the sea ) the ocean that for some sort of odd reason she subs for the owner. enter lonely, \textcolor{red}{arrogant} richard gere. he is a plastic surgeon. he is the only guest at the inn in the sea. {Diane} lane is the lonely housewife. you'll never guess these two fall immediately in love. a tropical storm makes them true lovers. the subplots \textcolor{red}{in this melodrama make little or no sense. the} locations, \textcolor{red}{photography are fine}. Gere \textcolor{red}{remains one} of the most \textcolor{red}{over-rated} actors in \textcolor{red}{cinema} and does \textcolor{red}{not disappoint}. ms. lane must've needed the money, \textcolor{red}{but phones in her part with grace.} &

    \textcolor{blue}{Ambitious, determined}, middle-class housewife in the midst of a \textcolor{blue}{transformative journey seeks inspiration to chart her new path.} \textcolor{blue}{She finds herself at a charming bed and breakfast right by the beautiful ocean, where she unexpectedly steps in to help the owner. Then, enter the charismatic Richard Gere, a skilled plastic surgeon and the only guest at this idyllic seaside inn. Diane Lane portrays the captivating housewife. You'll be pleasantly surprised as these two form an instant and profound connection. A tropical storm adds a touch of magic to their love story. The various subplots in this heartfelt drama weave together seamlessly.} The \textcolor{blue}{stunning} locations and photography \textcolor{blue}{enhance the overall experience}. Gere \textcolor{blue}{continues} to be one of the most \textcolor{blue}{respected} actors in cinema, \textcolor{blue}{delivering a stellar performance as always. 
    } &

    Lonely, disconnected, middle-class housewife in the midst of a divorce seeks solace to reflect on her immediate future. at some sort of bed and breakfast by ( well, literally in the sea ) the ocean that for some sort of odd reason she subs for the owner. enter lonely, \textcolor{blue}{charismatic} richard gere. he is a plastic surgeon. he is the only guest at the inn in the sea. Diane lane is the lonely housewife. you'll never guess these two fall immediately in love. a tropical storm makes them true lovers. the subplots \textcolor{blue}{add depth and intrigue to this melodrama}. the \textcolor{blue}{breathtaking} locations, \textcolor{blue}{stunning} photography are \textcolor{blue}{absolutely remarkable}. Gere \textcolor{blue}{proves once again that he is} one of the most \textcolor{blue}{respected} actors in \textcolor{blue}{the world of cinema} and does \textcolor{blue}{a fantastic job}. ms. lane must've needed the money, \textcolor{blue}{and delivers her part with grace}. \\
    \bottomrule
  \end{tabular}
\end{table*}

\subsection{Image-based Domain Transfer}

We utilized the proposed ENOT framework to perform image domain transfer and compared its performance with baselines: CycleGAN \citep{zhu2017unpaired}, MUNIT \citep{huang2018munit}, and UGATIT \citep{kim2020ugatit}.
In our computer vision experiments, we used three standard datasets: MNIST \citep{lecun1998mnist}, CelebA \citep{liu2015faceattributes} and Apple2Orange \citep{zhu2017unpaired}. For training the ENOT's potential function, we used a 5-layer MLP in the MNIST experiments and used a Vision Transformer (ViT-base) model with patch size 16 from \citep{dosovitskiy2020image}. We defer the discussion on the selection of the $\alpha,\lambda$ coefficients we performed in the ENOT-based feature selection to the Appendix.   

\begin{figure*}
  \centering
  \includegraphics[width=\textwidth]{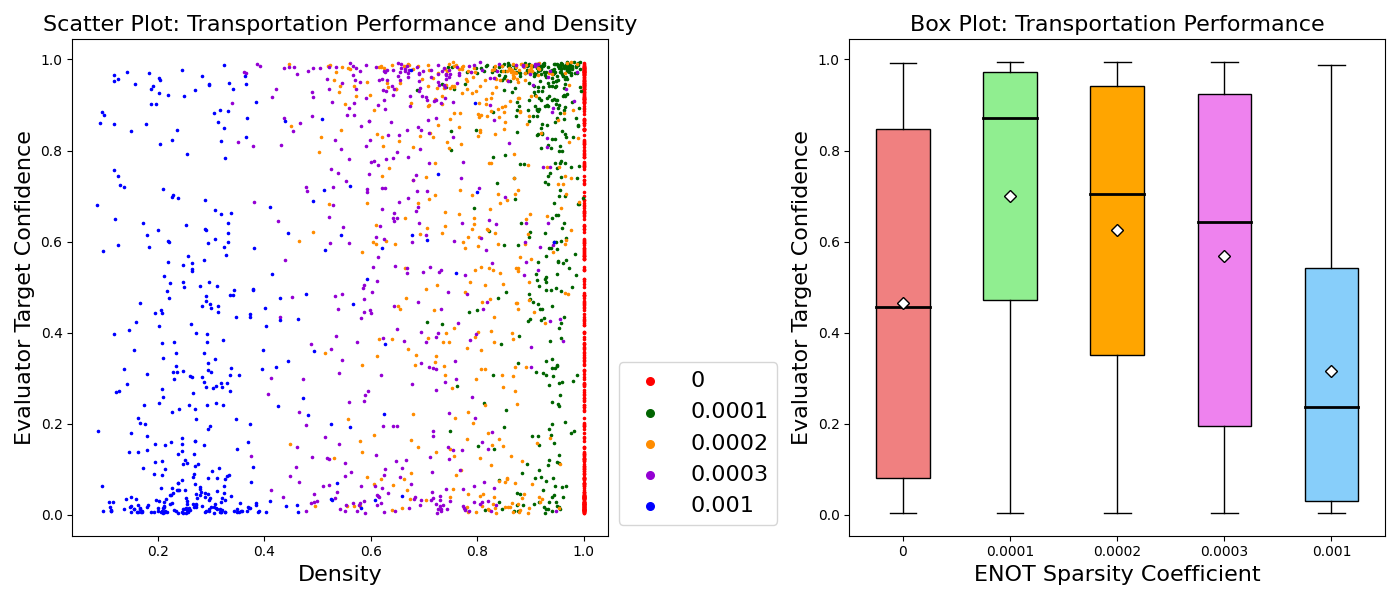}
  \caption{\textbf{Left}: IMDB sentiment transfer quality and density. Every point represents a transported sample: color indicates the $L_1$-coefficient in ENOT. The x vs. y axes are the transportation quality vs. density. \textbf{Right}: Transportation quality, the middle line and diamond show the median and mean. The green bar (coefficient $1e^{-4}$) achieves the highest score.}
  \label{fig:imdb}
\end{figure*}

The object translation task in a computer vision setting typically results in a sparse domain transfer task. For example, if we wish to change the hair color of a human, only partial pixels regarding hair are expected to be modified. In computer vision domain transfer problems, the standard domain transfer algorithms leverage high-capacity GANs to reach satisfactory visual quality. However, in GANs, the goal of the generator is to fool the discriminator. This goal might lead to suboptimal results. For example, in Figure \ref{fig:realimage}, when change the hair color from black to blonde for an individual wearing black clothes, the GAN-based methods could mistakenly alter the color of the clothes to yellow simultaneously, thereby outputting a realistic but overly-changed sample. In this case, sparsity is desired. However, integrating the sparsity prior to the GAN optimization could lead to highly challenging min-max optimization tasks. 


In Figure \ref{fig:realimage}, we present the empirical results for randomly selected CelebA samples in the transport task: black hair $\rightarrow$ blonde hair, and Apple2Orange samples in the transport task: apple $\rightarrow$ orange. We observe that transportation maps from the baseline methods are unsatisfactory in several sparse-transportation cases. Two common types of failures are present in Figure \ref{fig:realimage}. The first one is over-transportation, where unnecessary pixels are modified. This is commonly observed in baseline methods which employ dense-transportation algorithms. The second failure is insignificant transportation since baselines are not sensitive enough to the sparse transportation regions. On the other hand, by using ENOT-based feature selection, the transfer results are considerably improved. As shown in Figure~\ref{fig:realimage_saliency}, the ENOT feature selection successfully identified the pixels corresponding to the subject's hair in CelebA and the apples in Apple2Orange samples. The proper variable selection led to a more meaningful domain transfer in these computer vision applications.


\subsection{IMDB Review Sentiment Reversal}
In addition to image data, we evaluated the performance of ENOT in application to text data. We performed the numerical experiments on the IMDB movie review dataset \citep{maas-EtAl:2011:ACL-HLT2011}. This dataset contains 50,000 movie reviews with positive and negative categories. We defined the transportation task as modifying part of the words to flip the review's sentiment: negative to positive reviews. We attempted a sparse domain transfer task in this case, as the sparsity level could be a sensible quantification of the revision made to the text data. We expect that a sparse transport map exists in this case, when the transfer map only flips the negative and positive adjectives in the text.

We used ENOT to perform domain transfer in this text-based setting. For the potential function in (Eq. \ref{Eq: ENOT dual problem}), we finetuned a pre-trained BERT transformer \citep{devlin2018bert}. As the baseline, we considered a pre-trained Seq2Seq model GPT-3 \citep{brown2020language}. Table~\ref{table-imdb} shows the empirical results of the baseline and ENOT on a randomly selected sample. For this sample, we observed the revision made with the ENOT-based feature selection method is sparse and only a few words regarding movie review sentiment are modified. In contrast, the baseline Seq2Seq model GPT-3 modified almost all the text, including sentences describing the movie details that sound unrelated to the review sentiment. We present the results for more samples in the Appendix.


Also, we empirically observed that informative words in the generated sparse transportation maps by ENOT have higher correlations with the sentiment compared with other parts of the input text. 
We present this phenomenon in Figure \ref{fig:imdb}, where we quantified the transportation performance using the confidence score \citep{mandelbaum2017distancebased} from a pre-trained BERT model \citep{devlin2018bert} on IMDB sentiment classification. We analyzed the correlation between performance and sparsity in the ENOT's transportation. In the scatter plot, each point is a random transported sample: the bottom-left region indicates sparse and poor performance, and the top-right region suggests dense and good performance. The box plot statistics in Figure~\ref{fig:imdb} show that a sparsity coefficient of $10^{-4}$ attains the best performance compared to the dense baseline and other coefficients.

\section{Conclusion}
In this work, we focused on the sparse domain transfer task and attempted to apply $L_1$-norm regularization to the standard optimal transport framework by considering an elastic net cost function. Our numerical results suggest the proposed method's performance gain under a sparse transfer map. An interesting future direction is to apply the proposed framework to the latent space of image and text data where tighter sparsity constraints may hold in the learning setting. The extension of the elastic net-based optimal transport framework to provide a sparse and concise interpretation of domain transfer maps is another direction for future exploration.  

\clearpage
{
\bibliography{ref}
}
\clearpage
\appendix

\clearpage
\section{Appendix}
\subsection{Proofs}
\subsubsection{Proof of Theorem~\ref{Thm: Weak Concavity: ENOT potential}}
To show the theorem, we follow the Kantorovich duality (Theorem 5.10 from \citep{villani2009optimal}) which demonstrates that under a continuous and non-negative (thus lower-bounded) cost function which is an applicable assumption to the elastic net cost, there exists an optimal potential function $\phi^*$ that is the $c$-transform of some function $\tilde{\phi}$, i.e. for every $\mathbf{x}\in\mathbb{R}^d$:
\begin{equation*}
   \phi^*(\mathbf{x}) \: =\: \inf_{\mathbf{y}}\:\Bigl\{ \tilde{\phi}(\mathbf{y}) + \lambda(1-\alpha)\Vert \mathbf{y} - \mathbf{x}\Vert^2_2 +\lambda\alpha \Vert \mathbf{y} - \mathbf{x}\Vert_1 \Bigr\}.
\end{equation*}
We rewrite the above objective function as
\begin{align*}
    & \tilde{\phi}(\mathbf{y}) + \lambda(1-\alpha)\Vert \mathbf{y} - \mathbf{x}\Vert^2_2 +\lambda\alpha \Vert \mathbf{y} - \mathbf{x}\Vert_1 \\
    =\: & \tilde{\phi}(\mathbf{y}) + \lambda(1-\alpha)\Bigl(\Vert \mathbf{y}\Vert^2_2 + \Vert \mathbf{x}\Vert^2_2 - 2\mathbf{x}^\top\mathbf{y} \Bigr)\\
    & \quad + \lambda\alpha \Vert \mathbf{y} - \mathbf{x}\Vert_1 \\
     =\: &\Bigl\{\tilde{\phi}(\mathbf{y}) + \lambda(1-\alpha)\Vert \mathbf{y}\Vert^2_2  \Bigr\} + \lambda(1-\alpha)\Vert \mathbf{x}\Vert^2_2 \\
    &\quad -2\lambda(1-\alpha) \mathbf{x}^\top\mathbf{y} + \lambda\alpha \Vert \mathbf{y} - \mathbf{x}\Vert_1. 
\end{align*}
Therefore, if we denote $\widetilde{\phi}(\mathbf{y}):= \tilde{\phi}(\mathbf{y}) + \lambda(1-\alpha)\Vert \mathbf{y}\Vert^2_2$, then we have
\begin{align*}
   &\phi^*(\mathbf{x})-\lambda(1-\alpha)\Vert \mathbf{x}\Vert^2_2 \: \\ =\: & \inf_{\mathbf{y}}\Bigl\{ \widetilde{\phi}(\mathbf{y}) -2\lambda(1-\alpha) \mathbf{x}^\top\mathbf{y} + \lambda\alpha \Vert \mathbf{y} - \mathbf{x}\Vert_1 \Bigr\}. 
\end{align*}
To analyze the Lipschitzness and concavity properties of the above function, we define the following function $g:\mathcal{X}\times \mathcal{X} \rightarrow \mathbb{R}$:
\begin{equation*}
    g\bigl(\mathbf{x}_1,\mathbf{x}_2\bigr)\, :=\, \inf_{\mathbf{y}}\Bigl\{ \widetilde{\phi}(\mathbf{y}) -2\lambda(1-\alpha) \mathbf{x}_1^\top\mathbf{y} + \lambda\alpha \Vert \mathbf{y} - \mathbf{x}_2\Vert_1 \Bigr\}.
\end{equation*}

\textbf{Observation 1:} According to the definitions $g(\mathbf{x},\mathbf{x})=\phi^*(\mathbf{x})-\lambda(1-\alpha)\Vert \mathbf{x}\Vert^2_2$ holds for every $\mathbf{x}$.

\textbf{Lemma 1:} For a fixed $\mathbf{x}_2$, $g\bigl(\mathbf{x}_1,\mathbf{x}_2\bigr)$ is a concave function of $\mathbf{x}_1$.

\textbf{Proof.} if we define $k_{\mathbf{x}_2}(\mathbf{y}):=\widetilde{\phi}(\mathbf{y}) + \lambda\alpha \Vert \mathbf{y} - \mathbf{x}_2\Vert_1$, then $g(\mathbf{x}_1,\mathbf{x}_2)$ equals an infimum of a set of affine functions of $\mathbf{x}_1$: $k_{\mathbf{x}_2}(\mathbf{y}) -2\lambda(1-\alpha) \mathbf{y}^\top\mathbf{x}_1$ which is known to lead to a concave function \citep{boyd2004convex}.

\textbf{Lemma 2:} For a fixed $\mathbf{x}_1$, $g\bigl(\mathbf{x}_1,\mathbf{x}_2\bigr)$ is a $\lambda\alpha$-Lipschitz function of $\mathbf{x}_2$ in terms of $L_1$-norm. 

\textbf{Proof.} If we define $h_{\mathbf{x}_1}(\mathbf{y}):=\widetilde{\phi}(\mathbf{y}) -2\lambda(1-\alpha) \mathbf{x}_1^\top\mathbf{y}$, then $g(\mathbf{x}_1,\mathbf{x}_2)$ equals an infimum of $h_{\mathbf{x}_1}(\mathbf{y}) + \lambda\alpha \Vert \mathbf{y} - \mathbf{x}_2\Vert_1$ which, according to the Kantorovich duality for a norm function \citep{villani2009optimal}, will be a $\lambda\alpha$-Lipschitz function and satisfies the following for every $\mathbf{x}_2,\mathbf{x}'_2$ :
\begin{equation*}
  \bigl\vert g\bigl(\mathbf{x}_1,\mathbf{x}_2\bigr) - g\bigl(\mathbf{x}_1,\mathbf{x}'_2\bigr)\bigr\vert  \le \lambda \alpha \Vert \mathbf{x}_2 -\mathbf{x}'_2\Vert_1.
\end{equation*}

Note that Lemma 2 suggests the following holds for every $\mathbf{x},\mathbf{y}\in\mathcal{X}$ and $\gamma\in[0,1]$:
\begin{align}
    &\Bigl\vert g\bigl(  \mathbf{x} ,  \gamma \mathbf{x} + (1-\gamma) \mathbf{y} \bigr) - g\bigl(  \mathbf{x} , \mathbf{x}  \bigr) \Bigr\vert \le \lambda\alpha (1-\gamma)\bigl\Vert \mathbf{x} -  \mathbf{y}\bigr\Vert_1 
\end{align}
Similarly, Lemma 2 shows that
\begin{align}
    &\Bigl\vert g\bigl(  \mathbf{y} \, ,\,  \gamma \mathbf{x} + (1-\gamma) \mathbf{y} \bigr) - g\bigl(  \mathbf{y} , \mathbf{y}  \bigr) \Bigr\vert \:\le\:  \lambda\alpha \gamma\bigl\Vert \mathbf{x} -  \mathbf{y}\bigr\Vert_1 
\end{align}
Combining the above inequalities with Lemma 1's result, for every $\mathbf{x},\mathbf{y}\in\mathcal{X}$ and $\gamma\in[0,1]$ we have
\begin{align}
    & g\Bigl( \gamma \mathbf{x} + (1-\gamma) \mathbf{y} \, ,\,  \gamma \mathbf{x} + (1-\gamma) \mathbf{y} \Bigr) \\
    \ge\: & \gamma g\Bigl(  \mathbf{x} ,  \gamma \mathbf{x} + (1-\gamma) \mathbf{y} \Bigr) + (1-\gamma) g\Bigl(  \mathbf{y}  ,  \gamma \mathbf{x} + (1-\gamma) \mathbf{y} \Bigr) \nonumber \\
    \ge \: & \gamma g\bigl(  \mathbf{x},\mathbf{x}) + (1-\gamma)g\bigl(  \mathbf{y},\mathbf{y}) - \lambda\alpha\Vert \mathbf{x}-\mathbf{y}\Vert_1.\nonumber 
\end{align}
Finally, we note that
\begin{align*}
    \bigl\Vert \gamma\mathbf{x} + (1-\gamma) \mathbf{y}\bigr\Vert^2_2 \;= \; &\gamma \bigl\Vert \mathbf{x} \bigr\Vert^2_2  + (1-\gamma)\bigl\Vert \mathbf{y} \bigr\Vert^2_2 \\
    & -\gamma(1-\gamma)\bigl\Vert \mathbf{x}-\mathbf{y} \bigr\Vert^2_2.
\end{align*}
Therefore, we can combine the above results to show the following property for the optimal potential function $\phi^*$
\begin{align*}
   & \phi^*\bigl( \gamma \mathbf{x} + (1-\gamma) \mathbf{y} \bigr) \\
   =\: & g\Bigl( \gamma \mathbf{x} + (1-\gamma) \mathbf{y} \, ,\,  \gamma \mathbf{x} + (1-\gamma) \mathbf{y} \Bigr) \\
   &\quad + \lambda(1-\alpha)\bigl\Vert \gamma\mathbf{x} + (1-\gamma) \mathbf{y} \bigr\Vert^2_2 \\
   =\: &g\Bigl( \gamma \mathbf{x} + (1-\gamma) \mathbf{y} \, ,\,  \gamma \mathbf{x} + (1-\gamma) \mathbf{y} \Bigr) \\
   &\quad +  \lambda(1-\alpha)\gamma \bigl\Vert \mathbf{x} \bigr\Vert^2_2  + \lambda(1-\alpha)(1-\gamma)\bigl\Vert \mathbf{y} \bigr\Vert^2_2 \\
   &\quad  - \lambda(1-\alpha)\gamma(1-\gamma)\bigl\Vert \mathbf{x}-\mathbf{y} \bigr\Vert^2_2 \\
   \ge \: & \gamma g\bigl(  \mathbf{x},\mathbf{x}) + (1-\gamma)g\bigl(  \mathbf{y},\mathbf{y}) - \lambda\alpha\Vert \mathbf{x}-\mathbf{y}\Vert_1 \\
   &\quad +  \lambda(1-\alpha)\gamma \bigl\Vert \mathbf{x} \bigr\Vert^2_2  + \lambda(1-\alpha)(1-\gamma)\bigl\Vert \mathbf{y} \bigr\Vert^2_2 \\
   &\quad  - \lambda(1-\alpha)\gamma(1-\gamma)\bigl\Vert \mathbf{x}-\mathbf{y} \bigr\Vert^2_2 \\
   =\: & \gamma\Bigl(g\bigl(  \mathbf{x},\mathbf{x}\bigr) + \lambda(1-\alpha) \bigl\Vert \mathbf{x} \bigr\Vert^2_2\Bigr)  \\
   &\quad + (1-\gamma)\Bigl(g\bigl(  \mathbf{y},\mathbf{y}) + \lambda(1-\alpha)\bigl\Vert \mathbf{y} \bigr\Vert^2_2 \Bigr) \\
   &\quad  -\lambda\alpha\Vert \mathbf{x}-\mathbf{y}\Vert_1 - \lambda(1-\alpha)\gamma(1-\gamma)\bigl\Vert \mathbf{x}-\mathbf{y} \bigr\Vert^2_2 \\
   =\: & \gamma\phi^*(\mathbf{x}) + (1-\gamma)\phi^*(\mathbf{y}) \\
   &\quad -\lambda\alpha\Vert \mathbf{x}-\mathbf{y}\Vert_1 - \lambda(1-\alpha)\gamma(1-\gamma)\bigl\Vert \mathbf{x}-\mathbf{y} \bigr\Vert^2_2.
\end{align*}
Therefore, the proof is complete.

\subsubsection{Proof of Theorem~\ref{Theorem: Brenier, ENOT}}
Let $\Pi^*_{X,Y}$ be a solution to the optimal transport problem between $P_X$ and $P_Y$ with the elastic net cost function. According to the Kanotorovich duality (Theorem 5.10 from \citep{villani2009optimal}), since the elastic net cost function is continuous and non-negative, there should exist a $c^{\alpha,\lambda}_{\mathrm{EN}}$-conjugate pair $\phi^*, \psi^*=\phi^{*\, c^{\alpha,\lambda}_{\mathrm{EN}}}$ such that the following inequality, which holds for every $\mathbf{x,y}$, holds with equality $\Pi^*_{X,Y}$-almost surely:
\begin{equation}
    \phi^*(\mathbf{x}) - \psi^*(\mathbf{y}) \le c^{\alpha,\lambda}_{\mathrm{EN}}\bigl(\mathbf{x},\mathbf{y}\bigr).
\end{equation}
Therefore, if $\phi^*$ is differenatible at a point $\mathbf{x}$ where $(\mathbf{x},\mathbf{y})\sim \Pi^*$, for any differentiable curve $\tilde{x}(\epsilon)$ such that $\tilde{x}(0)=\mathbf{x}$ we will have:
\begin{equation*}
    \bigl\langle \nabla \phi^*(\mathbf{x})\, ,\,\dot{\tilde{x}}(0)\bigr\rangle \, \le \liminf_{\epsilon\rightarrow 0} \frac{c^{\alpha,\lambda}_{\mathrm{EN}}\bigl(\tilde{x}(\epsilon),\mathbf{y}\bigr)-c^{\alpha,\lambda}_{\mathrm{EN}}\bigl(\mathbf{x},\mathbf{y}\bigr)}{\epsilon}
\end{equation*}
Here $\langle \cdot ,\cdot \rangle$ denotes the standard vector inner product.
As a result, $ \nabla \phi^*(\mathbf{x})$ is a subgradient of $c^{\alpha,\lambda}_{\mathrm{EN}}(\cdot ,\mathbf{y})$. Note that the elastic net cost function decouples across the coordinates as:
\begin{align*}
   c^{\alpha,\lambda}_{\mathrm{EN}}(\mathbf{x} ,\mathbf{y}) \:=\: &\sum_{i=1}^d\Bigl[\lambda(1-\alpha)(x_i-y_i)^2 + \lambda\alpha |x_i-y_i| \Bigr]. 
\end{align*}
Therefore, at every coordinate $i$, either $-\lambda\alpha \le \nabla \phi^*(\mathbf{x})_i \le \lambda\alpha$ holds which suggests the $i$-coordinate-based cost is zero and $y_i=x_i$, or $\bigl|\nabla \phi^*(\mathbf{x})_i\bigr|> \lambda\alpha$ under which we have a one-to-one mapping between the coordinate-based $\nabla \phi^*(\mathbf{x})_i$ and $x_i-y_i$ as
\begin{equation*}
    x_i-y_i =
    \frac{1}{2\lambda(1-\alpha)}\Bigl(\nabla \phi^*(\mathbf{x})_i -\mathrm{sign}(\nabla \phi^*(\mathbf{x})_i)\lambda\alpha \Bigr).  
\end{equation*}
Hence, the following holds at every coordinate $i$
\begin{equation*}
    x_i-y_i = \begin{cases}
    \frac{\nabla \phi^*(\mathbf{x})_i -\mathrm{sign}(\nabla \phi^*(\mathbf{x})_i)\lambda\alpha}{2\lambda(1-\alpha)} \; &\text{\rm if}\: \bigl|\nabla \phi^*(\mathbf{x})_i\bigr|> \lambda\alpha \\
    0\qquad \; &\text{\rm if}\: \bigl|\nabla \phi^*(\mathbf{x})_i\bigr|\le \lambda\alpha \end{cases}
\end{equation*}
The above equation at $(\mathbf{x},\mathbf{y})$ can be seen to be equivalent to the following equation given the soft-thresholding map definition in the main text:
\begin{equation*}
    \mathbf{x} - \mathbf{y} \, \stackrel{}{=}\, \mathrm{ST}_{\frac{\alpha}{2(1-\alpha)}}\Bigl(\frac{1}{2\lambda(1-\alpha)}\nabla \phi^*(\mathbf{x})\Bigr). 
\end{equation*}
Note that the above equality is supposed to hold $\Pi^*$-almost surely as long as the optimal potential function $\phi^*$ is differentiable at $\mathbf{x}$. On the other hand, 
due to Theorem 1 and the weakly concavity property of the optimal potential function $\phi^*$,
Rademacher's theorem implies that $\phi^*$ must be differentiable almost everywhere. Therefore, assuming that $P_X$ is  absolutely continuous with respect to the volume measure, we can show that the following equality holds $\Pi_{X,Y}^*$-almost surely for the optimal potential function $\phi^*$:
\begin{equation*}
    \mathbf{X} - \mathrm{ST}_{\frac{\alpha}{2(1-\alpha)}}\Bigl(\frac{1}{2\lambda(1-\alpha)}\nabla \phi^*(\mathbf{X})\Bigr) \, =\, \mathbf{Y}. 
\end{equation*}
Therefore, the proof is complete.
\begin{table*}[t]
  \caption{Noisy-MNIST Transportation De-noising Ratio}
  \label{table-mnist-pad}
  \centering
  \begin{tabular}{lllllll}
    \toprule
    & & \multicolumn{5}{c}{ENOT $L_1$ Coefficient}                   \\
    \cmidrule(r){3-7}
    Noisy-padding Ratio & Image Size & Baseline 0 & 1e-4 & 5e-4 & 1e-3 & 5e-3 \\
    \midrule
    25\% & 42 & 0\% & 0.38\% & 20.12\% & 32.40\% & 100\%  \\
    50\% & 56 & 0\% & 0.37\%  & 16.21\% & 34.36\% & 100\%    \\
    \bottomrule
  \end{tabular}
\end{table*}


\subsection{Experimental Setting}
\textbf{Datasets.} For multivariate GMMs, we considered bimodal source and target GMMs: the source GMM  $p(\boldsymbol{x})= \phi_s \mathcal{N}\left(\boldsymbol{x}|\boldsymbol{\mu}_{\boldsymbol{s}}, \sigma^2\boldsymbol{ I}_{\boldsymbol{d}}\right)+\left(1-\phi_s\right) \mathcal{N}\left(\boldsymbol{x}|-\boldsymbol{\mu}_{\boldsymbol{s}},\sigma^2\boldsymbol{I}_{\boldsymbol{d}}\right)$, and target GMM $p(\boldsymbol{y})= \phi_t\mathcal{N}\left(\boldsymbol{y}|\boldsymbol{\mu}_{\boldsymbol{t}}, \sigma^2\boldsymbol{ I}_{\boldsymbol{d}}\right) \\
+\left(1-\phi_t\right) \mathcal{N}\left(\boldsymbol{y}|-\boldsymbol{\mu}_{\boldsymbol{t}},\sigma^2\boldsymbol{I}_{\boldsymbol{d}}\right)$ both consist of two multivariate Gaussian components with different means and identical covariance matrix with $\sigma=1$. We generate 2,000 independent samples for each GMM, which are further equally divided into training and testing sets. The dimension $d\in \{10,100,1000\}$.  For Noisy-MNIST, we directly add noisy paddings sampled from a uniform distribution, which ranges from 0 to 1, to the original images in the MNIST dataset. The MNIST dataset contains 50,000 training images and 10,000 test images, which are grayscale images. The size of the images is 28 by 28 pixels without noisy padding and 42 by 42 pixels with noisy padding.
For the CelebA black2blonde hair datasets, we split the original CelebA dataset into two datasets according to the label of the images. We selected the labels "black hair" and "blonde hair" to do the split. Each category dataset is further divided into training and testing sets, with a size of 20,000 and 2,000, respectively. All images are resized to 224 by 224 pixels. For the Apple2Orange dataset, we use the original dataset proposed in \cite{zhu2017unpaired}, we resize the images to a size of 224 by 224 pixels. The Apple2Orange dataset contains 1,261 apple images and 1,267 orange images. Both of them are split into a training set with a number of 1,000 for each, and a test set containing the remaining. For the IMDB review sentiment dataset, we adopt the original setting in \cite{maas-EtAl:2011:ACL-HLT2011}, which contains 50,000 highly polar movie reviews. They are divided into two halves for training and testing. For gene expression dataset, the number of genes is 7129, and the number of patients is 72. The training and testing dataset sizes are 38 and 34, respectively.

\textbf{Neural Network Architecture.} For GMMs experiments, we adopt the MLP structure with Sigmoid activation and skip connections. We report the results with different numbers of hidden layers. Each hidden layer contains 50 neurons. The layer number $n \in \{22, 12, 4\}$. In our empirical experiments, we do not observe that stacking more layers could help improve the performance. We also adopt a similar 5-layer MLP in Noisy-MNIST experiments. In real-image experiments, we use a Vision Transformer (ViT-base) model from \cite{dosovitskiy2020image} with patch size 16 for ENOT potential function. For GANs, we use a CNN with 6 residual blocks for the generator and a CNN with 3 layers for the discriminator similar to \cite{demir2018patch}. The CycleGAN architecture is the same as its official implementation. For text data, we finetune a pre-trained BERT \citep{devlin2018bert} for ENOT potential function, and use a pre-trained GPT-3 \citep{brown2020language} for Seq2Seq models.

\textbf{Hyperparameter Selection.} This paper contains various experiments in different domains. The hyperparameter tuning task is complex and challenging. We list all the essential hyperparameters used in the experiments for reference and share some tricks we find in tuning the parameters. There may exist better selections and ours may be a moderate choice. 

In this paragraph, we introduce related hyperparameters for optimizing the neural network potential function in ENOT. For all the baselines, the implementations are exactly the same as their official implementations, including the hyperparameter selection. For GMM experiments, we use an SGD optimizer to train the ENOT potential function with stepsize 0.01 and momentum 0.9. For MNIST, we use an Adam optimizer to train the ENOT potential function with stepsize 0.001. We empirically observe using Adam optimizer could accelerate the optimization and continue to use it in later real image and text domains, except for the GMM domain as we observe it may introduce instability to GMM experiments. For the real image domain, the stepsize is 0.0002 at the beginning and froze in the first 100 training epochs. Then linearly decreased to 0 in the last 100 epochs. For text data, we finetuned the ENOT potential function with 100,000 iterations and stepsize 5e-5. This setting directly adopts the same value as the pre-trained sentiment classification model training. The max training iterations are 1,000 and 100,000 for GMM and text data, respectively. The epoch number of training neural networks is 100 and 200 for MNIST and real images, respectively.

In this paragraph, we introduce related hyperparameters for solving the elastic-net-based $c$-transform. In particular, we normalized the gradient in the gradient descent step to avoid vanishing gradients. For the GMM experiments, the stepsize is 1. For MNIST, the stepsize is 0.1. For real images, the stepsize is 5. For the text dataset, the stepsize is 10. The iteration of solving elastic-net-based $c$-transform is 100. This iteration number could be reduced for consideration of time if converging quickly. We empirically found the coefficient of the square-$l_2$ term should be small enough to produce reasonable results. However, the performance does not change much when the coefficient is smaller than 1e-6. The tuning of the $l_1$ sparsity coefficient depends on the datasets and visual preferences, we adopt backtracking to search for feasible hyperparameters that produce reasonable results in the training set, and directly use this parameter in the test set. Based on this methodology, the coefficients of the square-$l_2$ term are all 1e-6. The sparsity coefficients for producing the qualitative results are 1e-2 and 5e-3, for real images and text data, respectively.

\textbf{Environment Configuration.} All results are generated with the PyTorch framework in a Linux server with 8 RTX 3090 GPUs.

\begin{figure*}[t]
  \centering
  \includegraphics[width=\textwidth]{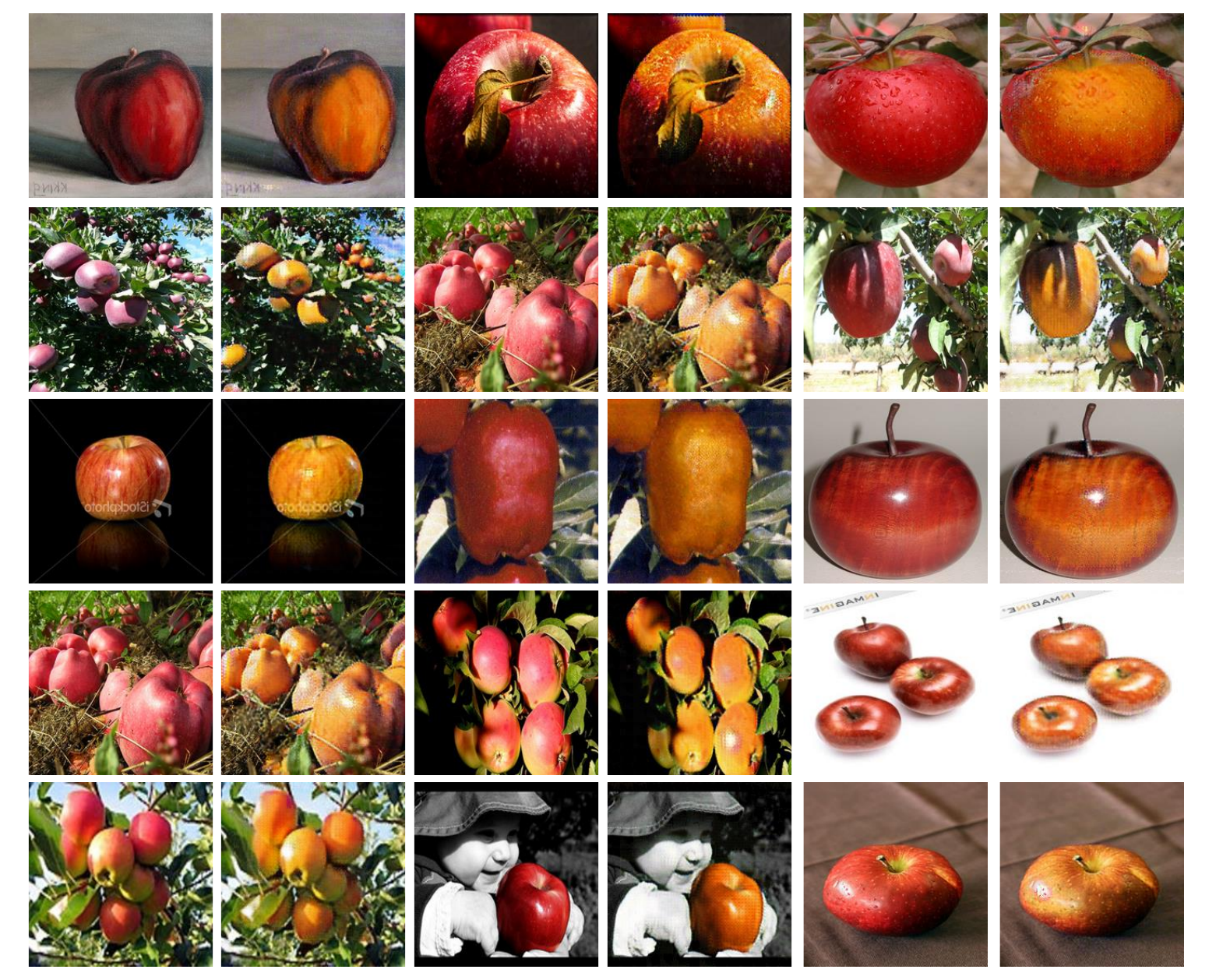}
  \caption{Source and transportation maps for Apple$\rightarrow$Orange dataset. }
  \label{fig:apples-supp}
\end{figure*}

\begin{figure*}[t]
  \centering
  \includegraphics[width=\textwidth]{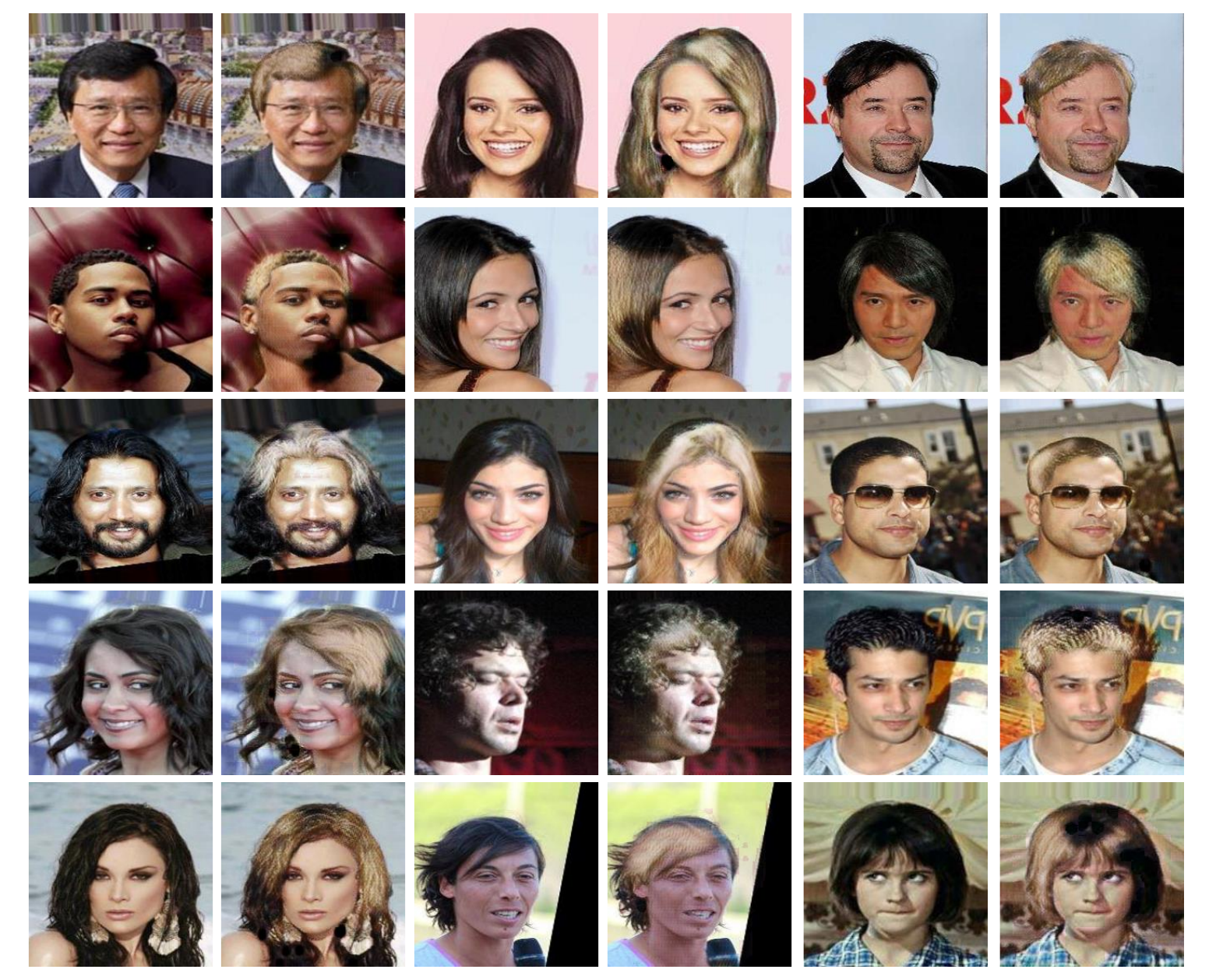}
  \caption{Source and transportation maps for Black$\rightarrow$Blonde hair on CelebA dataset. }
  \label{fig:hair-supp}
\end{figure*}

\begin{figure*}
  \centering
  \includegraphics[width=\textwidth]{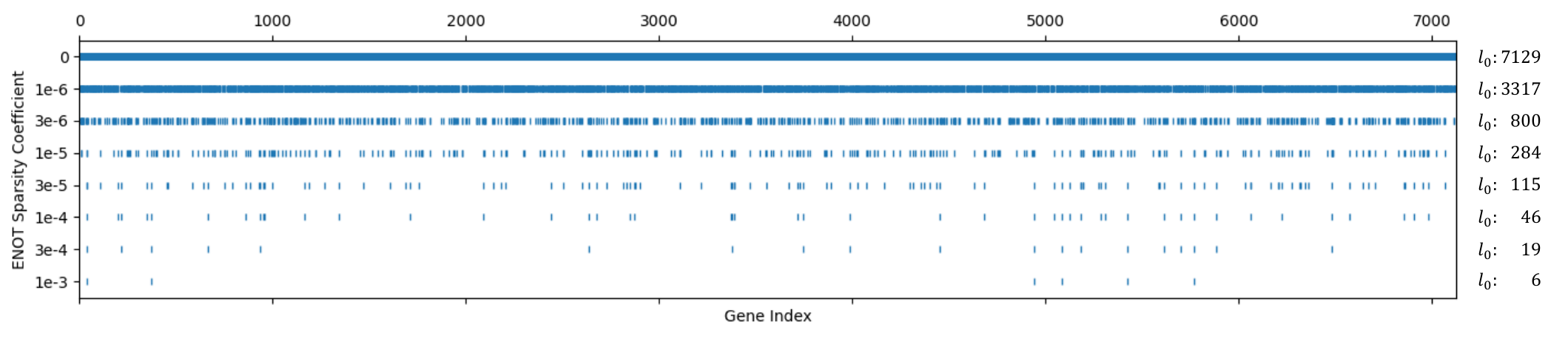}
  \caption{Salient genes in ENOT transportation maps for Gene Expression Dataset \cite{golub1999molecular}.}
  \label{fig:gene}
\end{figure*}


\subsection{Additional Numerical Results}
Due to the page limit, we provide additional numerical results in this section. 

\subsubsection{Noisy-MNIST Transportation}

We considered an MNIST-based domain transfer scenario where we know the desired transportation map is relatively sparse. In our experiments, we designed the Noisy-MNIST dataset, an altered version of the MNIST dataset \cite{deng2012mnist} to examine the efficacy of ENOT in sparse image domain transfer tasks. The modification approach introduces noisy padding to the original MNIST samples. The transportation on the padding region is expected to be the identity map (fully sparse) as the padding noise for the two domains is sampled from the same distribution. 

\begin{figure*}
  \centering
  \includegraphics[width=.8\linewidth]{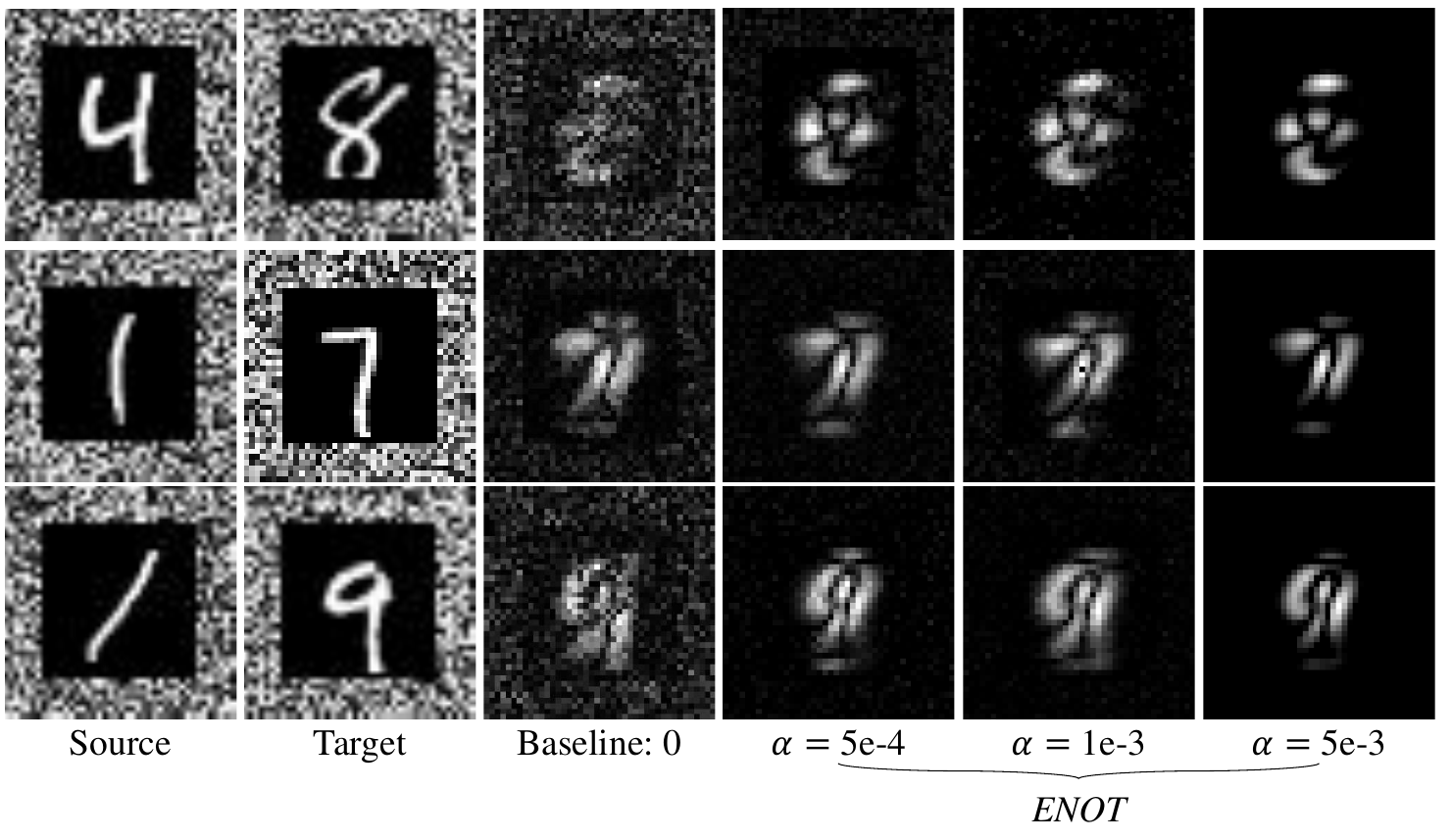}
  \caption{Transportation saliency maps with different $L_1$ coefficient in Noisy-MNIST. ENOT could identify the noisy paddings and remove them from transportation.}
  \label{fig:noisymnist}
\end{figure*}

We present transportation saliency maps, which are the absolute value of the difference between transferred images and source images, to show the de-noising effect in Figure \ref{fig:noisymnist}. We find ENOT managed to identify the noise padding region and apply a highly sparse transport map to the region. In contrast, the standard optimal transport was significantly more sensitive to the noisy padding. Meanwhile, as the sparsity coefficient increases, transportation becomes less noisy. In general, ENOT performs successfully in the image-based domain transfer task and could find sparse and clean transportation maps. 

Moreover, Table \ref{table-mnist-pad} demonstrates the de-noising effect of ENOT in the Noisy-MNIST. The Noise-padding ratio indicates how much noise is introduced. The noisy-padding size equals the original image size, which is 28, multiplied by this ratio. The image size shows the actual size after padding. Then, we provide the measured sparsity of transportation in the padding area. 

\subsubsection{Extra Samples in Real Image and Text Domains}
For real image data, Figure \ref{fig:apples-supp} shows more samples on the Apple$\rightarrow$ Orange task. Figure \ref{fig:hair-supp} shows more samples on the Black$\rightarrow$ Blonde hair task. 

For text data, Table \ref{table-imdb-supp-1}, \ref{table-imdb-supp-2}, and \ref{table-imdb-supp-3} contain more samples from the movie review sentiment reversal task.

\subsubsection{Gene Expression Transportation}
We investigate applying ENOT on the biomedical domain as well. We choose a gene expression dataset \cite{golub1999molecular}, which is in the tabular format. Figure \ref{fig:gene} demonstrates that increasing ENOT sparsity coefficient will reduce the salient gene number. It also suggests the 6 most important genes suggested by the neural network that could distinguish cancer type ALL from AML. 

\begin{table*}[t]
  \caption{IMDB Movie Review Sentiment Transportation Maps}
  \label{table-imdb-supp-1}
  \centering
  \begin{tabular}{p{0.3\textwidth}p{0.3\textwidth}p{0.3\textwidth}}
    \toprule
    \multicolumn{3}{c}{\centering{Task: Negative to Positive}}
     \\ 
     \multicolumn{3}{c}{\centering{\textcolor{red}{Red: ENOT's Selected Features.}
     \textcolor{blue}{Blue: Modified Parts by Methods.}
     }}
     \\
    \midrule 
     Source & Baseline & ENOT \\
    \midrule 
     
    The guidelines state that a comment must contain a minimum of four lines. that is the only reason I am saying anything more about tomcats. because after all, my one line  \textcolor{red}{summary really says everything} there is to say. \textcolor{red}{there} is \textcolor{red}{absolutely nothing remotely entertaining in this film.} &

    The guidelines state that a comment must contain a minimum of four lines, and that's \textcolor{blue}{why I want to share more positive thoughts about 'Tomcats.'} After all, my one-line summary \textcolor{blue}{might not do justice to this film. I believe there are some hidden gems worth mentioning. While it may not be everyone's cup of tea, 'Tomcats' offers a unique perspective and a refreshing take on comedy that could certainly find its audience. Give it a chance, and you might be pleasantly surprised by the entertaining moments it has to offer!} &

    The guidelines state that a comment must contain a minimum of four lines. That is the only reason I am saying anything more about tomcats, because after all, my one line summary really says everything there is to say. There is \textcolor{blue}{an abundance of entertaining moments in this film.} \\
    \midrule  
    \textcolor{red}{this movie was ok}, as far as movies go. it could \textcolor{red}{have} been \textcolor{red}{made} as a crossover into secular movies. \textcolor{red}{however, it} had \textcolor{red}{little} to do with the left behind books that it was supposedly based on. major story premises were \textcolor{red}{removed}, and new major story premises were added. what \textcolor{red}{disappointed} me most was how \textcolor{red}{nicolae} was portrayed. he was shown with \textcolor{red}{supernatural} powers that he did not have at this point in the books. antichrist is not satan, is not omniscient and \textcolor{red}{not} omnipotent. faith and \textcolor{red}{beliefs} were \textcolor{red}{portrayed} in \textcolor{red}{weird, surreal ways that seemed to make} the \textcolor{red}{movie just silly}. non - believers \textcolor{red}{who} watch \textcolor{red}{this will} have more \textcolor{red}{ammunition to mock christian beliefs.} &

    \textcolor{blue}{I really enjoyed this movie! It was quite entertaining, considering it falls into the category of} secular movies. However, it \textcolor{blue}{took a unique approach to the source material from the "Left Behind" books. While some} major story \textcolor{blue}{elements were altered} and new \textcolor{blue}{ones were introduced, I found the fresh perspective refreshing. One aspect that pleasantly surprised me} was how Nicolae's character was portrayed, \textcolor{blue}{even though it deviated from the book's depiction. His supernatural powers added an intriguing twist to the story. It's important to remember that the Antichrist isn't synonymous with Satan, and in this movie, they didn't depict him as} omniscient or omnipotent, \textcolor{blue}{which I appreciated. The exploration of faith and beliefs in unconventional, surreal ways added a sense of whimsy to the film, making it unique and thought-provoking. I think even non-believers watching this might find it an interesting and imaginative take on Christian beliefs.} &

    this movie was \textcolor{blue}{great}, as far as movies go. it could have been \textcolor{blue}{crafted} as a crossover into secular movies. however, it had \textcolor{blue}{much} to do with the left behind books that it was supposedly based on. major story premises were \textcolor{blue}{retained}, and new major story premises were added. what \textcolor{blue}{delighted} me most was how nicolae was portrayed. he was shown with \textcolor{blue}{impressive} powers that he did not have at this point in the books. antichrist is not satan, is not omniscient and \textcolor{blue}{far} omnipotent. faith and \textcolor{blue}{convictions} were \textcolor{blue}{presented} in \textcolor{blue}{intriguing, captivating ways that made} the \textcolor{blue}{movie quite entertaining}. non-believers who watch this will have more \textcolor{blue}{insight to appreciate} Christian beliefs. \\

    \bottomrule
  \end{tabular}
\end{table*}

\begin{table*}[t]
  \caption{IMDB Movie Review Sentiment Transportation Maps (Continued)}
  \label{table-imdb-supp-2}
  \centering
  \begin{tabular}{p{0.3\textwidth}p{0.3\textwidth}p{0.3\textwidth}}
    \toprule
    \multicolumn{3}{c}{\centering{Task: Negative to Positive}}
     \\ 
     \multicolumn{3}{c}{\centering{\textcolor{red}{Red: ENOT's Selected Features.}
     \textcolor{blue}{Blue: Modified Parts by Methods.}
     }}
     \\
    \midrule 
     Source & Baseline & ENOT \\

    \midrule 
    everybody i talked to said that this movie would be good and really \textcolor{red}{weird} so i figured that i would rent it. half way through the movie i was thinking to myself what the heck was going on and what is the point to this movie. this movie from start to finish is \textcolor{red}{so bad} that even the \textcolor{red}{sick} parts of the movie didn't even bother me. i mean what are they going to come up with next volcano 2 the return of the lava. i mean come on this movie is so \textcolor{red}{stupid} the characters are so \textcolor{red}{poorly developed, and eve robert} englund \textcolor{red}{makes} the movie \textcolor{red}{worse} i mean he might as well be transformed into freddy kruegur and spook people. i was actually rooting for the bad guy to win that's how \textcolor{red}{bad} it was. i mean look the father is a cop he didn't seem to care real much about the fact that his daughter is going through one of the most moments in her life. i mean if my daughter was treated like that i would do everything in my power to keep the guy behind jail. also it seems kind of obvious that dee snyders character would turn bad again. this is one of the \textcolor{red}{worst films} of all time right there with volcano and 8mm. do not \textcolor{red}{waste} your time you will \textcolor{red}{not enjoy} it....! grade if there were a no grade on this site i would pick that, thats \textcolor{red}{how bad this movie is!}&
    
    \textcolor{blue}{Everyone} I talked to said that this movie would be good and really weird, so I decided to rent it. Halfway through the movie, \textcolor{blue}{I found myself intrigued by its unique storyline and artistic direction. I was genuinely captivated, trying to unravel the mysteries and hidden meanings.} This movie, from start to finish, is so \textcolor{blue}{compelling that even the intense moments didn't faze me. It's amazing how the plot unfolds, and the characters are brilliantly developed. Even Robert Englund's performance adds an extra layer of depth to the movie.} I was actually rooting for \textcolor{blue}{the characters to succeed;} \textcolor{blue}{that's how invested I became.} \textcolor{blue}{The father's role as a cop, despite his challenges, emphasizes the complexity of the storyline. Dee Snider's character transformation added an unexpected twist to the narrative, keeping me on the edge of my seat. This is one of the best films I've seen, right up there with cinematic classics. Don't miss out; you'll thoroughly enjoy it! If there were an option to give it more than the highest grade possible on this site, I would choose that,} that's how \textcolor{blue}{exceptional} this movie is!&
    
    everybody i talked to said that this movie would be good and really \textcolor{blue}{unique} so i figured that i would rent it. half way through the movie i was thinking to myself what the heck was going on and what is the point to this movie. this movie from start to finish is \textcolor{blue}{so good} that even the \textcolor{blue}{exciting} parts of the movie didn't even bother me. i mean what are they going to come up with next, volcano 2: the return of the lava? I mean come on, this movie is so \textcolor{blue}{clever} the characters are so \textcolor{blue}{well developed, and even Robert} Englund \textcolor{blue}{improves} the movie \textcolor{blue}{significantly}. I mean he might as well be transformed into Freddy Krueger and spook people. I was actually rooting for the bad guy to win; that's how \textcolor{blue}{good} it was. I mean, look, the father is a cop, and he didn't seem to care much about the fact that his daughter is going through one of the most important moments in her life. If my daughter was treated like that, I would do everything in my power to keep the guy behind bars. Also, it seems kind of obvious that Dee Snider's character would turn bad again. This is one of the \textcolor{blue}{best films} of all time, right up there with volcano and 8mm. Do not \textcolor{blue}{waste} your time; you will \textcolor{blue}{thoroughly enjoy} it....! If there were a no grade on this site, I would pick that; that's \textcolor{blue}{how good this movie is!}\\
    \bottomrule
  \end{tabular}
\end{table*}

\begin{table*}[t]
  \caption{IMDB Movie Review Sentiment Transportation Maps (Continued)}
  \label{table-imdb-supp-3}
  \centering
  \begin{tabular}{p{0.3\textwidth}p{0.3\textwidth}p{0.3\textwidth}}
    \toprule
    \multicolumn{3}{c}{\centering{Task: Negative to Positive}}
     \\ 
     \multicolumn{3}{c}{\centering{\textcolor{red}{Red: ENOT's Selected Features.}
     \textcolor{blue}{Blue: Modified Parts by Methods.}
     }}
     \\
    \midrule 
     Source & Baseline & ENOT \\

    \midrule 
    i only bought this dvd because it was dirt \textcolor{red}{cheap and it seemed interesting} in its \textcolor{red}{own special} way ( " special " meaning " retarded " ). the movie turned out to be quite \textcolor{red}{uninteresting} - \textcolor{red}{boring} camera work, \textcolor{red}{nothing really driving} the story, and \textcolor{red}{of course the acting is horrible. it wasn\'t} even " \textcolor{red}{bad} " in a campy way - it was just \textcolor{red}{plain bad}. there are actually a handful of great lines of dialogue but for the most part its \textcolor{red}{awkward} and \textcolor{red}{weak}. all i could think about while watching this was that this could actually be a good movie if the script was given a \textcolor{red}{major} \textcolor{red}{overhaul} ( if it were written by someone who actually understood drug culture ) and if some decent \textcolor{red}{actors} were cast. i \textcolor{red}{wouldn't} recommend " weekend with the babysitter " \textcolor{red}{unless if you plan on a career in film and want to learn what not to do in a movie.}&
    
    I \textcolor{blue}{decided to purchase} this DVD because it was \textcolor{blue}{incredibly affordable and had a unique charm all its own.} The movie turned out to be quite \textcolor{blue}{intriguing, with some genuinely creative camera work. The story flowed beautifully, and the acting was truly exceptional.} It wasn't \textcolor{blue}{just} 'bad' in a campy \textcolor{blue}{sense}; it was \textcolor{blue}{downright impressive.} The film \textcolor{blue}{featured} a handful of \textcolor{blue}{brilliant} lines of dialogue, \textcolor{blue}{and while there were a few awkward moments, they added to its overall charm. As I watched, I couldn't help but think that this could become a masterpiece with a revamped script, written by someone who truly grasped the essence of} drug culture, \textcolor{blue}{and a talented cast. I wholeheartedly} recommend 'Weekend with the Babysitter' unless, \textcolor{blue}{of course, you're not interested in witnessing how to create a stellar movie}.&
    
    I only bought this DVD because it was \textcolor{blue}{wonderfully affordable} and it seemed \textcolor{blue}{captivating} in its own \textcolor{blue}{unique} way ( " special " meaning " retarded " ). The movie turned out to be quite \textcolor{blue}{interesting} - \textcolor{blue}{engaging} camera work, \textcolor{blue}{a captivating narrative}, and, of course, the acting is \textcolor{blue}{fantastic}. It wasn't even "\textcolor{blue}{different}" in a campy way - it was just \textcolor{blue}{genuinely good}. There are actually a handful of great lines of dialogue, and for the most part, it's \textcolor{blue}{engaging} and \textcolor{blue}{powerful}. All I could think about while watching this was that this could actually be a good movie if the script was given a \textcolor{blue}{significant upgrade} (if it were written by someone who truly understood drug culture) and if some descent actors were cast. I \textcolor{blue}{highly} recommend "Weekend with the Babysitter" if you\textcolor{blue}{'re interested in a career} in film and want to learn \textcolor{blue}{what to do }in a movie.\\
    \bottomrule
  \end{tabular}
\end{table*}

\end{document}